%% file: main.tex
\documentclass[10pt,twocolumn,letterpaper]{article}

\usepackage{cvpr_2023}

\usepackage{graphicx}
\usepackage{amsmath}
\usepackage{amsthm}
\usepackage{amssymb}
\usepackage{booktabs}
\usepackage{enumitem}
\newcommand{\std}[1]{\scriptsize{$\pm$ #1}}

\usepackage[pagebackref,breaklinks,colorlinks]{hyperref}

\usepackage[capitalize]{cleveref}
\crefname{section}{Sec.}{Secs.}
\Crefname{section}{Section}{Sections}
\Crefname{table}{Table}{Tables}
\crefname{table}{Tab.}{Tabs.}
\usepackage[accsupp]{axessibility}

\def\cvprPaperID{10345}
\def\confName{CVPR}
\def\confYear{2023}

\newtheorem{prop}{Proposition}
\usepackage{algorithm}
\usepackage{algorithmic}
\usepackage[accsupp]{axessibility}  

\title{DART: Diversify-Aggregate-Repeat Training\\Improves Generalization of Neural Networks}

\author{%
 Samyak Jain \thanks{Equal Contribution.  $^\mp$ Equal contribution second authors.  Correspondence to Samyak Jain $<$samyakjain.cse18@itbhu.ac.in$>$, Sravanti Addepalli $<$sravantia@iisc.ac.in$>$. $^\diamond$ Indian Institute of Technology, Varanasi  \qquad $^\S$ Indian Institute of Technology, Dhanbad. $^\ddagger$ Work done during internship at Vision and AI Lab, Indian Institute of Science, Bangalore.} ~  $^\diamond$ $^\ddagger$   \quad Sravanti Addepalli \footnotemark[1]  \\\qquad Pawan Kumar Sahu $^\mp$ $^\S$ $^\ddagger$  \quad Priyam Dey $^\mp$ \quad R.Venkatesh Babu  \\ 
 Vision and AI Lab, Indian Institute of Science, Bangalore}

\begin{document}
\maketitle
\begin{abstract}

Generalization of Neural Networks is crucial for deploying them safely in the real world. Common training strategies to improve generalization involve the use of data augmentations, ensembling and model averaging. In this work, we first establish a surprisingly simple but strong benchmark for generalization which utilizes diverse augmentations within a training minibatch, and show that this can learn a more balanced distribution of features. Further, we propose Diversify-Aggregate-Repeat Training (DART) strategy that first trains diverse models using different augmentations (or domains) to explore the loss basin, and further Aggregates their weights to combine their expertise and obtain improved generalization. 
We find that Repeating the step of Aggregation throughout training improves the overall optimization trajectory and also ensures that the individual models have sufficiently low loss barrier to obtain improved generalization on combining them. We theoretically justify the proposed approach and show that it indeed generalizes better. In addition to improvements in In-Domain generalization, we demonstrate SOTA performance on the Domain Generalization benchmarks in the popular DomainBed framework as well. Our method is generic and can easily be integrated with several base training algorithms to achieve performance gains. Our code is available here: \url{https://github.com/val-iisc/DART}.

\end{abstract}
\vspace{-0.5cm}
\section{Introduction}
\label{sec:intro}

Deep Neural Networks have outperformed classical methods in several fields and applications owing to their remarkable generalization. Classical Machine Learning theory assumes that test data is sampled from the same distribution as train data. This is referred to as the problem of In-Domain (ID) generalization \cite{foret2020sharpness, jiang2019fantastic, dziugaite2017computing, petzka2021relative, huang2020understanding}, where the goal of the model is to generalize to samples within same domain as the train dataset. This is often considered to be one of the most important requirements and criteria to evaluate models. However, in several cases, the test distribution may be different from the train distribution. For example, surveillance systems are expected to work well at all times of the day, under different lighting conditions and when there are occlusions, although it may not be possible to train models using data from all these distributions. It is thus crucial to train models that are robust to distribution shifts, i.e., with better Out-of-Domain (OOD) Generalization \cite{hendrycks2019benchmarking}.  In this work, we consider the problems of In-Domain generalization and Out-of-Domain Generalization of Deep Networks. For the latter, we consider the popular setting of Domain Generalization \cite{Li_2018_ECCV, gulrajani2020search, cha2021swad}, where the training data is composed of several source domains and the goal is to generalize to an unseen target domain. 

The problem of generalization is closely related to the Simplicity Bias of Neural Networks, due to which models have a tendency to rely on simpler features that are often spurious correlations to the labels, when compared to the harder robust features \cite{shah2020pitfalls}. For example, models tend to rely on weak features such as background, rather than more robust features such as shape, causing a drop in object classification accuracy when background changes \cite{geirhos2018imagenet,xiao2020noise}. A common strategy to alleviate this is to use data augmentations \cite{cubuk2018autoaugment, yun2019cutmix,cubuk2020randaugment,zhang2017mixup,devries2017improved, lim2021noisy, batchaug, rame2021mixmo} or data from several domains during training \cite{gulrajani2020search}, which can result in invariance to several spurious correlations, improving the generalization of models. Shen et al. \cite{shen2022data} show that data augmentations enable the model to give higher importance to harder-to-learn robust features by delaying the learning of spurious features. We extend their observation by showing that training on a combination of several augmentation strategies (which we refer to as \textit{Mixed} augmentation) can result in the learning of a balanced distribution of diverse features. Using this, we obtain a strong benchmark for ID generalization as shown in Table-\ref{table:motivation}. However, as shown in prior works \cite{addepalli2022efficient}, the impact of augmentations in training is limited by the capacity of the network in being able to generalize well to the diverse augmented data distribution. Therefore, increasing the diversity of training data demands the use of larger model capacities to achieve optimal performance. This demand for higher model capacity can be mitigated by training specialists on each kind of augmentation and ensembling their outputs \cite{lakshminarayanan2017simple, dietterich2000ensemble,  saurabhensemble, zhangenseble}, which results in improved performance as shown in Table-\ref{table:motivation}.
Another generic strategy that is known to improve generalization is model-weight averaging  \cite{izmailov2018averaging, wortsman21alearningsubspace, wortsman2022model}, which results in a flatter minima.

\begin{figure}
\centering
        \includegraphics[width=0.9\linewidth]{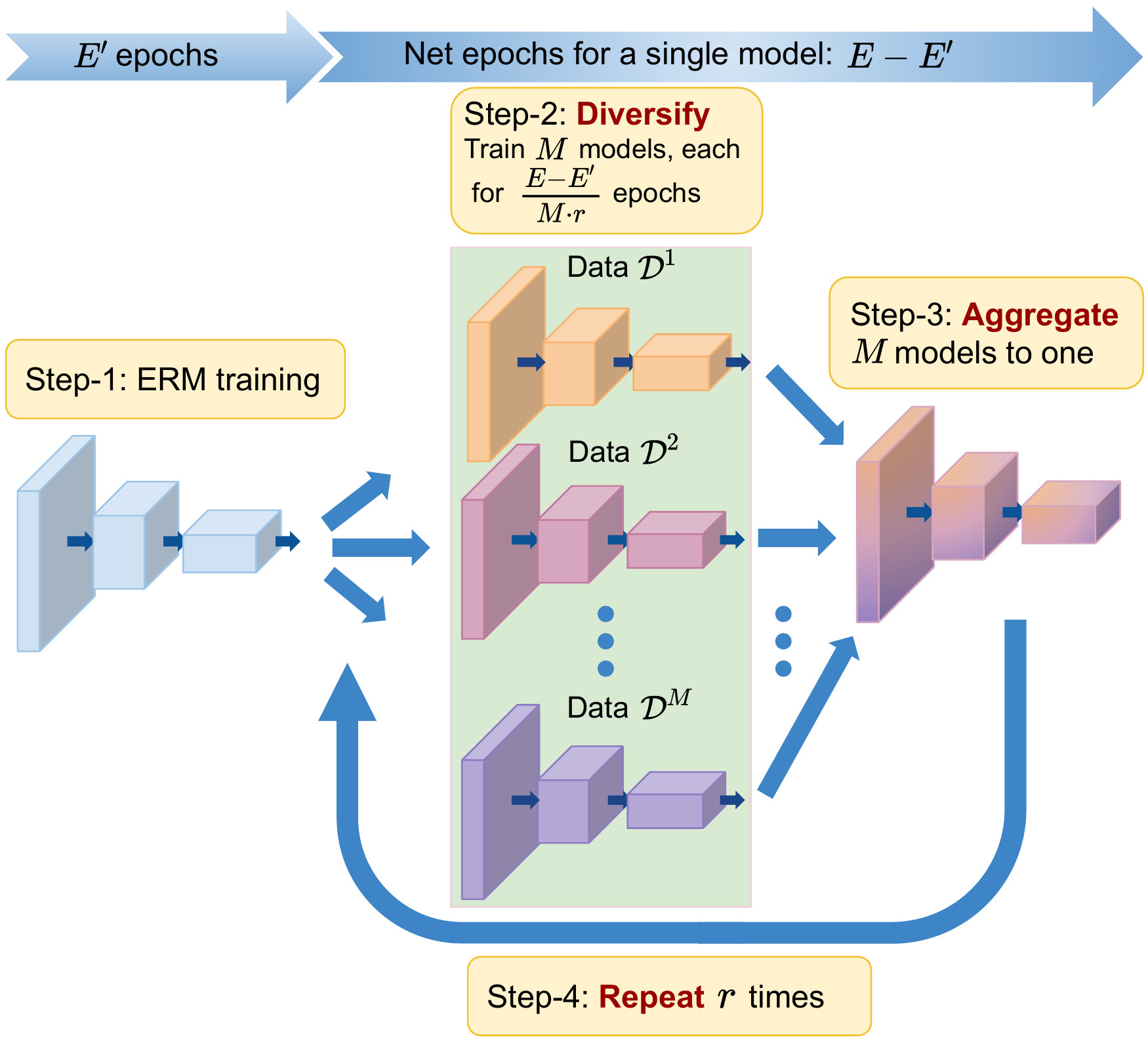}
        \caption{Schematic Diagram of the proposed method DART}
        \label{fig:dart}
        \vspace{-0.3cm}
\end{figure}

In this work, we aim to combine the benefits of the three strategies discussed above - diversification, specialization and model weight averaging, while also overcoming their individual shortcomings. We propose a \textbf{D}iversify-\textbf{A}ggregate-\textbf{R}epeat \textbf{T}raining strategy dubbed DART (Fig.\ref{fig:dart}), that first trains $M$ \emph{Diverse} models after a few epochs of common training, and then \emph{Aggregates} their weights to obtain a single generalized solution. The aggregated model is then used to reinitialize the $M$ models which are further trained post aggregation. This process is \emph{Repeated} over training to obtain improved generalization. The \emph{Diversify} step allows models to explore the loss basin and specialize on a fixed set of features. The \emph{Aggregate} (or Model Interpolation) step robustly combines these models, increasing the diversity of represented features while also suppressing spurious correlations. Repeating the \emph{Diversify-Aggregate} steps over training ensures that the $M$ diverse models remain in the same basin thereby permitting a fruitful combination of their weights. We justify our approach theoretically and empirically, and show that intermediate model aggregation also increases the learning time for spurious features, improving generalization. We present our key contributions below: 

\input{tables/motivation.tex}

\begin{itemize}
\vspace{-0.2cm}
    \item We present a strong baseline termed Mixed-Training (MT) that uses a combination of diverse augmentations for different images in a training minibatch.
    \vspace{-0.2cm}
    \item We propose a novel algorithm DART, that learns specialized diverse models and aggregates their weights iteratively to improve generalization.
    \vspace{-0.2cm}
    \item We justify our method theoretically, and empirically on several In-Domain (CIFAR-10, CIFAR-100, ImageNet) and Domain Generalization (OfficeHome, PACS, VLCS, TerraIncognita, DomainNet) datasets. 
\end{itemize}

\vspace{-0.3cm}
\section{Background: Mode Connectivity of Models}
\vspace{-0.15cm}
\label{sec:background}
The overparameterization of Deep networks leads to the existence of multiple optimal solutions to any given loss function \cite{keskar2016large,zhang2021understanding,neyshabur2017exploring}. Prior works \cite{garipov2018loss, draxlericml, nguyen2019connected} have shown that all such solutions learned by SGD lie on a non-linear manifold, and are connected to each other by a path of low loss. Frankle \etal \cite{frankle2020linear} further showed that converged models that share a common initial optimization path are linearly connected with a low loss barrier. This is referred to as the \emph{linear mode connectivity} between the models. Several optimal solutions that are linearly connected to each other are said to belong to a common \emph{basin} which is separated from other regions of the loss landscape with a higher \emph{loss barrier}. Loss barrier between any two models $\theta_1$ and $\theta_2$ is defined as the maximum loss attained by the models, $\hat{\theta} = \alpha \cdot \theta_1 + (1 - \alpha) \cdot \theta_2 ~~~ \forall ~~~ \alpha \in [0,1]$. 

The linear mode connectivity of models facilitates the averaging of weights of different models in a common basin resulting in further gains. In this work, we leverage the linear mode connectivity of diverse models trained from a common initialization to improve generalization. 
\vspace{-0.1cm}
\section{Related Works}
\vspace{-0.1cm}
\label{sec:related}
\subsection{Generalization of Deep Networks}
\vspace{-0.1cm}
Prior works aim to improve the generalization of Deep Networks by imposing invariances to several factors of variation. This is achieved by using data augmentations during training \cite{cubuk2018autoaugment, devries2017improved, yun2019cutmix, zhang2017mixup, cubuk2020randaugment, verma2019manifold, lim2021noisy, hendrycks2020augmix}, or by training on a combination of multiple domains in the Domain Generalization (DG) setting \cite{Li_2017_ICCV, Li_2018_ECCV, hu2020domain, ilse2020diva, chuang2020estimating}. In DG, several works have focused on utilizing domain-specific features \cite{8053784, bui2021exploiting}, while others try to disentangle the features as domain-specific and domain-invariant for better generalization \cite{chattopadhyay2020learning, Li_2017_ICCV, 10.1007/978-3-642-33718-5_12, piratla2020efficient, unknown}. Data augmentation has also been exploited for Domain Generalization \cite{wang2020heterogeneous, Volpi_2019_ICCV, shi2020towards, qiao2020learning, volpi2018generalizing, shankar2018generalizing, xu2020robust, Yue_2019_ICCV, zhou2021domain, mancini2020towards, zhou2020learning} in order to increase the diversity of training data and simulate domain shift. 
Foret \etal \cite{foret2020sharpness} show that minimizing the maximum loss within an $\ell_2$ norm ball of weights can result in a flatter minima thereby improving generalization. Gulrajani \etal \cite{gulrajani2020search} show that the simple strategy of ERM training on data from several source domains can indeed prove to be a very strong baseline for Domain Generalization. The authors also release DomainBed - which benchmarks several existing methods on some common datasets representing different types of distribution shifts. Recently, Cha \etal \cite{cha2022miro} propose MIRO, which introduces a Mutual-Information based regularizer to retain the superior generalization of the pre-trained initialization or Oracle, thereby demonstrating significant improvements on DG datasets. The proposed method DART achieves SOTA on the popular DG benchmarks and shows further improvements when used in conjunction with several other methods (Table-\ref{table:dg_2}) ascribing to its orthogonal nature. 

\vspace{-0.05cm}
\subsection{Averaging model weights across training} 
\vspace{-0.05cm}
Recent works have shown that converging to a flatter minima can lead to improved generalization \cite{foret2020sharpness, jiang2019fantastic, dziugaite2017computing, petzka2021relative, huang2020understanding, stutz2021relating}. Exponential Moving Average (EMA) \cite{polyak1992acceleration} and Stochastic Weight Averaging (SWA) \cite{izmailov2018averaging} are often used to average the model weights across different training epochs so that the resulting model converges to a flatter minima, thus improving generalization at no extra training cost.  
Cha \etal \cite{cha2021swad} theoretically show that converging to a flatter minima results in a smaller domain generalization gap. The authors propose SWAD that overcomes the limitations of SWA in the Domain Generalization setting and combines several models in the optimal solution basin to obtain a flatter minima with better generalization. 
We demonstrate that our approach effectively integrates with EMA and SWAD for In-Domain and Domain Generalization settings respectively to obtain further performance gains (Tables-\ref{table:main_tab}, \ref{table:dg_cut}). 

\vspace{-0.05cm}
\subsection{Averaging weights of fine-tuned models} 
\vspace{-0.05cm}
While earlier works combined models generated from the same optimization trajectory, Tatro \etal \cite{tatro2020optimizing} showed that for any two converged models with different random initializations, one can find a permutation of one of the models so that fine-tuning the interpolation of this with the second model leads to improved generalization. On a similar note, Zhao \etal \cite{zhao2020bridging} proposed to achieve robustness to backdoor attacks by fine-tuning the linear interpolation of pre-trained models. More recently, Wortsman \etal \cite{wortsman2022model} proposed Model Soups and showed that in a transfer learning setup, fine-tuning and then averaging different models with same pre-trained initialization but with different hyperparameters such as learning rates, optimizers and augmentations can improve the generalization of the resulting model. The authors further note that this works best when the pre-trained model is trained on a large heterogeneous dataset. While all these approaches work only in a fine-tuning setting, the proposed method incorporates the interpolation of differently trained models in the regime of \textit{training from scratch}, allowing the learning of models for longer schedules and larger learning rates. 
\vspace{-0.05cm}
\subsection{Averaging weights of differently trained models} 
\vspace{-0.05cm}
Wortsman \etal \cite{wortsman21alearningsubspace} propose to average the weights of multiple models trained simultaneously with different random initializations by considering the loss of a combined model for optimization, while performing gradient updates on the individual models. Additionally, they minimize the cosine similarity between model weights to ensure that the models learned are diverse. While this training formulation does learn diverse connected models, it leads to individual models having sub-optimal accuracy (Table-\ref{table:main_tab}) since their loss is not optimized directly. DART overcomes such issues since the individual models are trained directly to optimize their respective classification losses. Moreover, the step of intermediate interpolation ensures that the individual models also have better performance when compared to the baseline of standard ERM training on the respective augmentations (Fig.\ref{fig:gains_last} in the Supplementary).

\begin{figure}
\centering
        \includegraphics[width=0.8\linewidth]{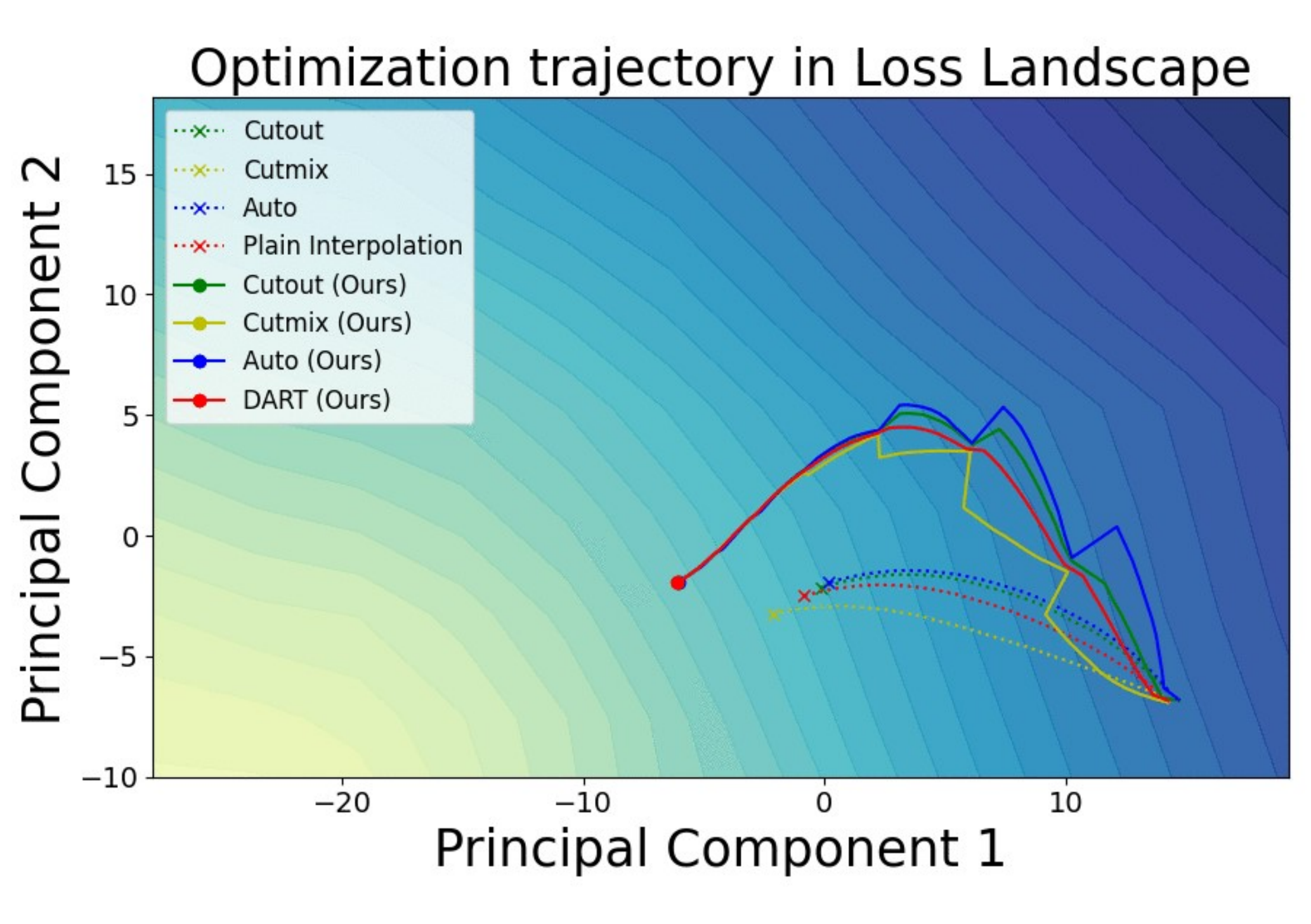}
         \vspace{-0.4cm}
        \caption{\textbf{Optimization trajectory} of the proposed approach DART when compared to independent ERM training on each augmentation. Axes represent the top two PCA directions obtained using the weights of DART training. The initial common point on the right represents the model obtained after 100 epochs of Mixed Training (MT). The trajectory shown is for an additional 100 epochs, with a total training budget of 200 epochs.}
        \label{fig:trajectory}
        \vspace{-0.5cm}
\end{figure}

\section{Proposed Method: DART}
\input{algorithm/algo}

A series of observations from prior works \cite{garipov2018loss, draxlericml, nguyen2019connected,frankle2020linear} have led to the conjecture that models trained independently with different initializations could be linearly connected with a low loss barrier, when different permutations of their weights are considered, suggesting that \emph{all solutions effectively lie in a common basin} \cite{entezari2021role}. Motivated by these observations, we aim at designing an algorithm that explores the basin of solutions effectively with a robust optimization path and combines the expertise of several diverse models to obtain a single generalized solution. 

We show an outline of the proposed approach - \emph{Diversify-Aggregate-Repeat Training}, dubbed DART, in Fig.\ref{fig:dart}. Broadly, the proposed approach is implemented in four steps - i) ERM training for $E'$ epochs in the beginning, followed by ii) Training $M$ \emph{Diverse} models for $\lambda/M$ epochs each, iii) \emph{Aggregating} their weights, and finally iv) \emph{Repeating} the steps \emph{Diversify-Aggregate} for $E-E'$ epochs. 

A cosine learning rate schedule is used for training the model for a total of $E$ epochs with a maximum learning rate of $\mathrm{LR}_{max}$. We present the implementation of DART in Algorithm-\ref{alg:WATIG}, and discuss each step in detail below:

\begin{enumerate}
\vspace{-0.2cm}
    \item \textbf{Traversing to the \emph{Basin} of optimal solutions:} Since the goal of the proposed approach is to explore the \emph{basin} of optimal solutions, the first step is to traverse from a randomly initialized model upto the periphery of this basin. Towards this, the proposed \emph{Mixed-Training} strategy discussed in  Section-\ref{sec:intro} is performed on a combination of several augmentations $D^*$ for the initial $E'$ epochs (L4-L5 in Alg.\ref{alg:WATIG}). 
    \vspace{-0.2cm}
    \item \label{itm:diversify} \textbf{Diversify - Exploring the \emph{Basin}:} In this step, $M$ diverse models $f_{\theta^k}$ initialized from the Mixed-Training model (L8 in Alg.\ref{alg:WATIG}), are trained using the respective datasets $D^k$ (L10 in Alg.\ref{alg:WATIG}). These are generated using diverse augmentations in the In-Domain setting, and from a combination of different domains in the Domain Generalization setting. We set $|D^k| = |D|/M$ where $D$ is the original dataset.
    \vspace{-0.2cm}
    \item \textbf{Aggregate - Combining diverse experts:} Owing to the initial common training for $E'$ epochs, the $k$ diverse models lie in the same basin, enabling an effective aggregation of their weights using simple averaging (L12 in Alg.\ref{alg:WATIG}) to obtain a more generalized solution $\theta$. Aggregation is done after every $\lambda$ epochs. 
    \vspace{-0.2cm}
    \item \textbf{Repeat:} Next, all $k$ models are reinitialized using the common model $\theta$ (L13 of Alg.\ref{alg:WATIG}), after which the individual models are trained for $\lambda$ epochs on their respective datasets $D^k$ as discussed in Step-\ref{itm:diversify}, and the process continues for a total of $E-E'$ epochs. \\
    \vspace{-0.5cm}
\end{enumerate}

\noindent \textbf{Visualizing the Optimization Trajectory:} We compare the optimization trajectory of the proposed approach DART with independent training on the same augmentations in Fig.\ref{fig:trajectory} after a common training of $E'=100$ epochs on Mixed augmentations. The models explore more in the initial phase of training, and lesser thereafter, which is a result of the cosine learning rate schedule and reducing gradient magnitudes over training. The exploration in the initial phase helps in increasing the diversity of models, thereby improving the robustness to spurious features (as shown in Proposition-\ref{prop:Ours_noise}) leading to a better optimization trajectory, while the smaller steps towards the end help in retaining the flatter optima obtained after Aggregation. 
The process of repeated aggregation also ensures that the models remain close to each other, allowing longer training regimes. 

\vspace{-0.2cm}
\section{Theoretical Results}
\vspace{-0.1cm}
\label{theory_setup}
We use the theoretical setup from Shen \etal \cite{shen2022data} to show that the proposed approach DART achieves robustness to spurious features, thereby improving generalization.

\noindent \textbf{Preliminaries and Setup:}
We consider a binary classification problem with two classes $\{-1,1\}$. We assume that the dataset contains $n$ inputs and $K$ orthonormal robust features which are important for classification and are represented as $v_1, v_2, v_3, \dots, v_K$, in decreasing order of their frequency in the dataset. Let each input example $x$ be composed of two patches denoted as $(x_1, x_2) \in R^{d \times 2}$, where each patch is characterized as follows: i) \textbf{Feature patch}: $x_1=yv_{k^*}$ where $y$ is the target label of $x$ and $k^* \in [1,K]$, ii) \textbf{Noisy patch}: $x_2=\epsilon$ where $\epsilon \sim \mathcal{N}\left(0,\frac{\sigma^2}{d}I_d\right)$.

We consider a single layer convolutional neural network consisting of C channels, with $w=(w_1, w_2, w_3, \dots, w_C) \in R^{d \times C}$. The function learned by the neural network (F) is given by $F(w,x)=\sum\limits_{c=1}^C\sum\limits_{p=1}^2\phi(w_c, x_p)$, where $\phi$ is the activation function as defined by Shen \etal \cite{shen2022data}.

\vspace{0.1cm}
\noindent \textbf{Weights learned by an ERM trained model:}
Let $K_{cut}$ denote the number of robust features learned by the model. Following Shen \etal \cite{shen2022data}, we assume the learned weights to be a linear combination of the two types of features present in the dataset as shown below:
\begin{equation}
\label{eq:weight}
    w = \sum\limits_{k=1}^{K_{cut}}v_k + \sum\limits_{k>K_{cut}}y^{(k)}\epsilon^{(k)} 
\end{equation}

\noindent \textbf{Data Augmentations:} As defined by Shen \etal \cite{shen2022data}, an augmentation $T_k$ can be defined as follows ($K$ denotes the number of different robust patches in the dataset):
    \begin{multline}
    \label{eq:aug_transform}
\forall ~k'\in[1, K],~~\mathcal{T}_k(v_{k^{'}}) = v_{((k^{'} + k - 1) ~mod~K)+1}
    \end{multline}
    Assuming unique augmentations for each of the $m$ branches, the augmented data is defined as follows:
    \begin{multline}
    \label{eq:aug_union}
    D_{train}^{(aug)} = D_{train}~\cup~ \mathcal{T}_1(D_{train}) .. \cup~ \mathcal{T}_{m-1}(D_{train})
    \end{multline}
    where $D_{train}$ is the training dataset.
If $m=K$, each feature patch $v_i$ appears $n$ times in the dataset, thus making the distribution of all the feature patches uniform.

\vspace{0.1cm}
\noindent \textbf{Weight Averaging in DART:}
In the proposed method, we consider that $m$ models are being independently trained after which their weights are averaged as shown below: 
\vspace{-0.2cm}
\begin{equation}
\label{eq:weights_watig}
w =  \frac{1}{m}\sum\limits_{j=1}^{m}\sum\limits_{k=1}^{K_{{cut}_j}}v_{k_{j}} + \frac{1}{m}\sum\limits_{j=1}^{m}\sum\limits_{k>K_{{cut}_j}}y^{(k)}_j\epsilon^{(k)}_j
\end{equation}
Each branch is trained on the dataset $D_{train}^{(k)}$ defined as:
    \begin{equation}
    \label{eq:aug_def}
    D_{train}^{(k)} = \mathcal{T}_k(D_{train}), ~~ k\in[1,2,...,m]
    \end{equation}
    
\textbf{Propositions:} In the following propositions, we derive the convergence time for learning robust and noisy features, and compare the same with the bounds derived by Shen \etal \cite{shen2022data} in Section-\ref{sec:theory_analysis}. The proofs of all propositions are presented in Section-\ref{sec:theory} of the Supplementary. 
\\
\noindent \textbf{Notation:} Let $f_\theta$ denote a neural network obtained by averaging the weights of $m$ individual models $f_\theta^k$, $k \in [1,m]$ which are represented as shown in Eq.\ref{eq:weight}. $n$ is the total number of data samples in the original dataset $D_{train}$. $K$ is the number of orthonormal robust features in the dataset. The weights ${w_1, w_2, \dots, w_C}$ of each model $f_\theta^k$ are initialized as $w_c\sim \mathcal{N}\left(0,\sigma_{0}^2I_d\right) ~ \forall ~c \in [1,C]$, where C is the number of channels in a single layer of the model. $\frac{\sigma}{\sqrt{d}}$ is the standard deviation of the noise in noisy patches, $q$ is a hyperparameter used to define the activation (Details in Section-\ref{sec:theory} of the Supplementary), where $q\geq3$ and $d$ is the dimension of each feature patch and weight channel $w_c$.

\vspace{-0.15cm}
\begin{prop}
\label{prop:robust}
The convergence time for learning any feature patch $v_{i} ~~ \forall i\in[1,K]$ in at least one channel $c \in C$ of the weight averaged model  $f_\theta$ using the augmentations defined in Eq.\ref{eq:aug_def}, is given by $O\left(\frac{K}{\sigma_{0}^{q-2}}\right)$, if $\frac{\sigma^q}{\sqrt{d}} \ll \frac{1}{K}$, $m=K$.

\vspace{-0.2cm}
\end{prop}
\begin{prop}
\label{prop:SWA}
If the noise patches learned by each $f_\theta^k$ are $i.i.d.$ Gaussian random variables $\sim \mathcal{N}(0, \frac{\sigma^2}{d}I_d)$ then with high probability, convergence time of learning a noisy patch $\epsilon^{(j)}$ in at least one channels $c \in [1,C]$ of the weight averaged model $f_\theta$ is given by $O\left(\frac{nm}{\sigma_{0}^{q-2}\sigma^q}\right)$, if $d \gg n^2$.
\end{prop}
\vspace{-0.4cm}
\begin{prop}
\label{prop:Ours_noise}
If the noise learned by each $f_\theta^k$ are $i.i.d.$ Gaussian random variables $\sim \mathcal{N}\left(0, \frac{\sigma^2}{d}I_d\right)$, and model weight averaging is performed at epoch $T$, the convergence time of learning a noisy patch $\epsilon^{(j)}$ in at least one channels $c \in [1,C]$ of the weight averaged model $f_\theta$ is given by $T + O\left(\frac{nm^{(q-2)}d^{(q-2)/2}}{\sigma^{(2q-2)}}\right)$, 
if $d \gg n^2$.
\end{prop}

\section{Analysis on the Theoretical Results}
\label{sec:theory_analysis}
In this section, we present the implications of the theoretical results discussed above. While the setup in Section-\ref{theory_setup} discussed the existence of only two kinds of patches (feature and noisy), in practice, a combination of these two kinds of patches - termed as Spurious features - could also exist, whose convergence can be derived from the above results. 

\subsection{Learning Diverse Robust Features}

We first show that \emph{using sufficiently diverse data augmentations during training generates a uniform distribution of feature patches, encouraging the learning of diverse and robust features by the network}. We consider the use of $m$ unique augmentations in Eq.\ref{eq:aug_union} which transform each feature patch into a different one using a unique mapping as shown in Eq.\ref{eq:aug_transform}. The mapping in Eq.\ref{eq:aug_transform} can transform a skewed feature distribution to a more uniform distribution after performing augmentations. This results in $K_{cut}$ being sufficiently large in Eq.\ref{eq:weight}, which depends on the number of high frequency robust features, thereby encouraging the learning of a more balanced distribution of robust features. While Proposition-\ref{prop:robust} assumes that $m=K$, we show in Corollary \ref{prop-corollary} in the Supplementary that even when $m \neq K$, the learning of hard features is enhanced.

Shen \etal \cite{shen2022data} show that the time for learning any feature patch $v_k$ by at least one weight channel $c\in C$ is given by $O\left(\frac{1}{\sigma_{0}^{q-2}\rho_k}\right)$ if $\frac{\sigma^q}{\sqrt{d}} \ll \rho_k$, where $\rho_k$ is the fraction of the frequency of occurrence of feature patch $v_k$ divided by the total number of occurrences of all the feature patches in the dataset. The convergence time for learning feature patches is thus limited by the one that is least frequent in the input data. Therefore, by making the frequency of occurrence of all feature patches uniform, this convergence time reduces. In Proposition-\ref{prop:robust} we show that the same holds true even for the proposed method DART, where several branches are trained using diverse augmentations and their weights are finally averaged to obtain the final model. This justifies the improvements obtained in Mixed-Training (Eq.\ref{eq:weight}) and in the proposed approach DART (Eq.\ref{eq:weights_watig}) as shown in Table-\ref{table:main_tab}.

\subsection{Robustness to Noisy Features}

Firstly, the use of diverse augmentations in both Mixed-Training (MT) and DART results in better robustness to noisy features since the value of $K_{cut}$ in Eq.\ref{eq:weight} and Eq.\ref{eq:weights_watig} would be higher, resulting in the learning of more feature patches and suppressing the learning of noisy patches. \textit{The proposed method DART indeed suppresses the learning of noisy patches further, and also increases the convergence time for learning noisy features as shown in Proposition-\ref{prop:SWA}}. When the augmentations used in each of the $m$ individual branches of DART are diverse, the noise learned by each of them can be assumed to be $i.i.d.$ Under this assumption, averaging model weights at the end of training results in a reduction of noise variance, as shown in Eq.\ref{eq:weights_watig}. More formally, we show in Proposition-\ref{prop:SWA} that the \textit{convergence time of noisy patches increases by a factor of $m$ when compared to ERM training}. We note that this does not hold in the case of averaging model weights obtained during a single optimization trajectory as in SWA \cite{izmailov2018averaging}, EMA \cite{polyak1992acceleration} or SWAD \cite{cha2021swad}, since the noise learned by models that are close to each other in the optimization trajectory cannot be assumed to be $i.i.d.$

\subsection{Impact of Intermediate Interpolations}

We next analyse the impact of averaging the weights of the models at an intermediate epoch $T$ in addition to the interpolation at the end of training. The individual models are further reinitialized using the weights of the interpolated model as discussed in Algorithm-\ref{alg:WATIG}. As shown in Proposition-\ref{prop:Ours_noise}, averaging the weights of all branches at the intermediate epoch $T$ helps in increasing the convergence time of noisy patches by a factor $O\left(\frac{\sigma_0^{q-2}m^{q-3}d^{(q-2)/2}}{\sigma^{q-2}}\right)$ when compared to the case where models are interpolated only at the end of training as shown in Proposition-\ref{prop:SWA}. By assuming that $q>3$ and $d \gg n^2$ similar to Shen \etal \cite{shen2022data}, the lower bound on this can be written as $O\left(\frac{\sigma_0n}{\sigma}\right)$. We note that in a practical scenario this factor would be greater than 1, demonstrating the increase in convergence time for noisy patches when intermediate interpolation is done. 

\section{Experiments and Results}

In this section, we empirically demonstrate the performance gains obtained using the proposed approach DART on In-Domain (ID) and Domain Generalization (DG) datasets. We further attempt to understand the various factors that contribute to the success of DART. 

\textbf{Dataset Details:} To demonstrate In-Domain generalization, we present results on CIFAR-10 and CIFAR-100 \cite{Krizhevsky2009LearningML}, while for DG, we present results on the 5 real-world datasets on the DomainBed \cite{gulrajani2020search} benchmark - VLCS \cite{6751316}, PACS \cite{Li_2017_ICCV}, OfficeHome \cite{venkateswara2017deep}, Terra Incognita \cite{beery2018recognition} and DomainNet \cite{peng2019moment}, which represent several types of domain shifts with different levels of dataset and task complexities. 

\begin{figure*}
\centering
        \includegraphics[width=1\linewidth]{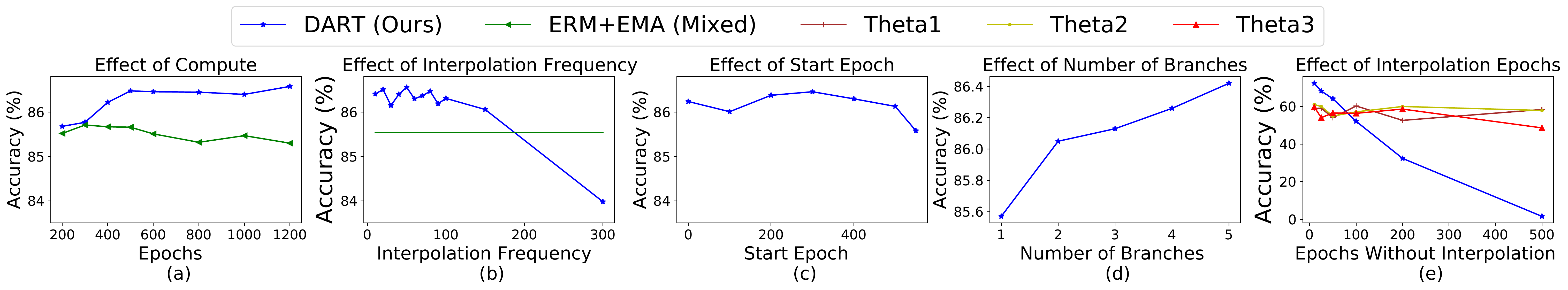}
         \vspace{-0.6cm}
        \caption{\textbf{Ablations on CIFAR-100, WideResNet-28-10}: (a-d) Experiments comparing DART with the Mixed-Training baseline using the standard training settings. (e) Varying the interpolation epoch after 50 epochs of common training using a fixed learning rate of 0.1.}
        \label{fig:num_models}
        \vspace{-0.2cm}
\end{figure*}

\textbf{Training Details (ID):} The training epochs are set to 600 for the In-Domain experiments on CIFAR-10 and CIFAR-100. To enable a fair comparison, the best performing configuration amongst 200, 400 and 600 total training epochs is used for the ERM baselines and Mixed-Training, since they may be prone to overfitting. We use SGD optimizer with momentum of 0.9, weight decay of 5e-4 and a cosine learning rate schedule with a maximum learning rate of 0.1. Interpolation frequency ($\lambda$) is set to 50 epochs for CIFAR-100 and 40 epochs for CIFAR-10. As shown in Fig-\ref{fig:num_models}(b), accuracy is stable when $\lambda \in [10,80]$. We present results on ResNet-18 and WideResNet-28-10 architectures.

\vspace{0.1cm}
\textbf{Training Details (DG):} Following the setting in DomainBed \cite{gulrajani2020search}, we use Adam \cite{kingma2014adam} optimizer with a fixed learning rate of 5e-5. The number of training iterations are set to 15k for DomainNet (due to its higher complexity) and 10k for all other datasets with the interpolation frequency being set to 1k iterations. ResNet-50\cite{7780459} was used as the backbone, initialized with Imagenet\cite{russakovsky2015imagenet} pre-trained weights. Best-model selection across training checkpoints was done based on validation results from the train domains itself, and no subset of the test domain was used. We use fixed values of hyperparameters for all datasets in the DG setting. As shown in Fig.\ref{fig:hyper_fig} (a) of the Supplementary, ID and OOD accuracies are correlated, showing that hyperparameter tuning based on ID validation accuracy as suggested by Gulrajani \etal \cite{gulrajani2020search} can indeed improve our results further. We present further details in Section-\ref{sec:train_details} of Supplementary. 
\input{tables/ID.tex}
\input{tables/Imagenet}
\input{tables/DG_cut}
\input{tables/DG_algos}

\textbf{In Domain (ID) Generalization:} In Table-\ref{table:main_tab}, we compare our method against ERM training with several augmentations, and also the strong Mixed-Training benchmark (MT) obtained by using either AutoAugment \cite{cubuk2018autoaugment}, Cutout \cite{devries2017improved} or Cutmix \cite{yun2019cutmix} for every image in the training minibatch uniformly at random. We use the same augmentations in DART as well, with each of the 3 branches being trained on one of the augmentations. As discussed in Section-\ref{sec:related}, the method proposed by Wortsman \etal \cite{wortsman21alearningsubspace} is closest to our approach, and hence we compare with it as well. We utilize Exponential Moving Averaging (EMA) \cite{polyak1992acceleration} of weights for the ERM baselines and the proposed approach for a fair comparison. On CIFAR-10, we observe gains of 0.19\% on using ERM-EMA (Mixed) and an additional 0.27\% on using DART. On CIFAR-100, 1.37\% improvement is observed with ERM-EMA (Mixed) and an additional 0.89\% with the proposed method DART. We also incorporate DART with SAM \cite{foret2020sharpness} and obtain $\sim 0.2\%$ gains over ERM + SAM with Mixed Augmentations as shown in Table-\ref{table:sam} of the Supplementary. The comparison of DART with the Mixed Training benchmark (ERM+EMA on mixed augmentations) on ImageNet-1K and fine-grained datasets, Stanford-Cars \cite{KrauseStarkDengFei-Fei_3DRR2013} and CUB-200 \cite{wah2011caltech} on an ImageNet pre-trained model is shown in Table-\ref{table:imagenet}. On ImageNet-1K,  we obtain 0.41\% gains on using RandAugment \cite{cubuk2020randaugment} across all the branches, and 0.14\% gains on using Pad-Crop, RandAugment and Cutout for different branches. We obtain gains of upto 1.5\% on fine-grained datasets.

\textbf{SOTA comparison - Domain Generalization:} We present results on the DomainBed \cite{gulrajani2020search} datasets in Table-\ref{table:dg_cut}. We compare only with ERM training (performed on data from a mix of all domains) and SWAD \cite{cha2021swad} in the main paper due to lack of space, and present a thorough comparison across all other baselines in Section-\ref{tabular:avg} of the Supplementary. For the DG experiments, we consider 4 branches ($M=4$), with 3 branches being specialists on a given domain and the fourth being trained on a combination of all domains in equal proportion. For the DomainNet dataset, we consider 6 branches due to the presence of more domains. On average, we obtain 2.8\% improvements over the ERM baseline without integrating with SWAD, and 1\% higher accuracy when compared to SWAD by integrating our approach with it. We further note from Table-\ref{table:dg_2} that the DART can be integrated with several base approaches - with and without SWAD, while obtaining substantial gains across the respective baselines. The proposed approach therefore is generic, and can be integrated effectively with several algorithms. As shown in the last row, we obtain substantial gains of 2.6\% on integrating DART with SWAD and a recent work MIRO \cite{cha2022miro} using CLIP initialization \cite{CLIP_paper} on a ViT-B/16 model \cite{dosovitskiy2020image}.

\input{tables/augs_impact.tex}
\textbf{Evaluation without imposing diversity across branches:} While the proposed approach imposes diversity across branches by using different augmentations, we show in Table-\ref{table:augs} that it works even without explicitly introducing diversity, by virtue of the randomness introduced by SGD and different ordering of input samples across models. We obtain an average improvement of 0.9\% over the respective baselines, and maximum improvement of 1.82\% using Cutout. This shows that the performance of DART is not dependent on data augmentations, although it achieves further improvements on using them.

\begin{figure}
\centering
        \includegraphics[width=1\linewidth]{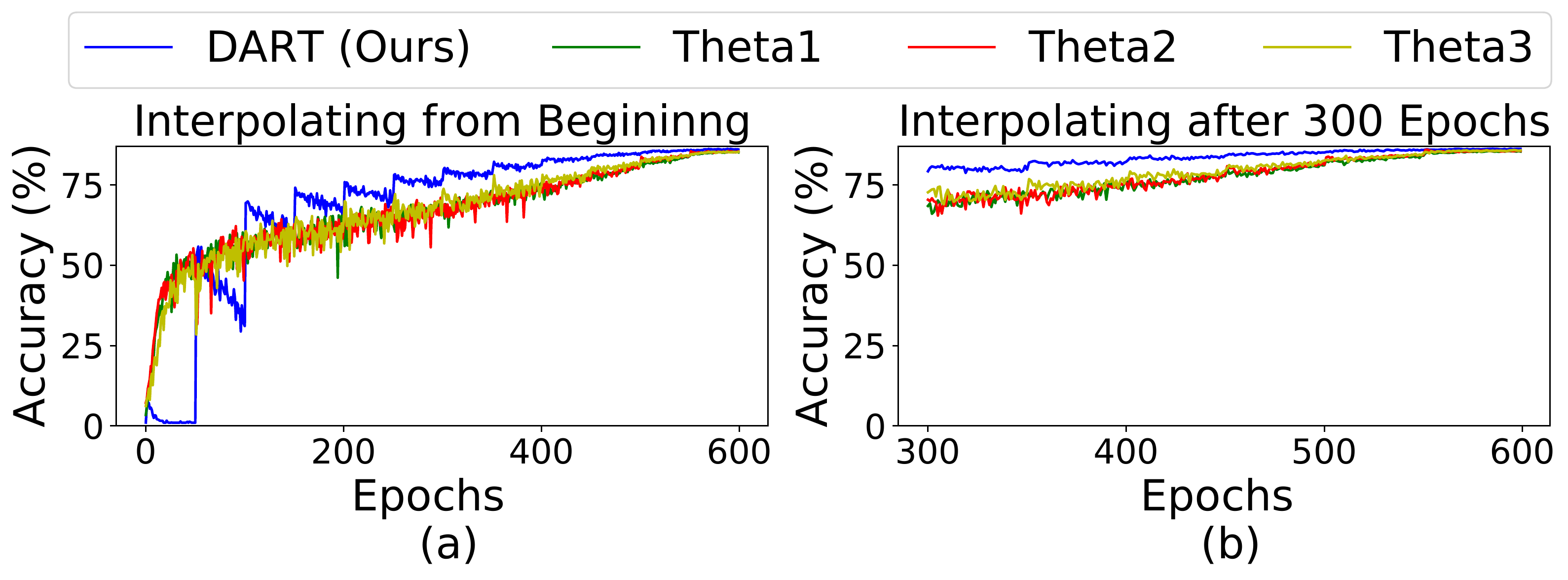}
         \vspace{-0.7cm}
         \caption{\textbf{Accuracy of DART across training epochs} for CIFAR-100 on WideResNet-28-10 model: Each branch is trained on different augmentations, whose accuracy is also plotted. Model Interpolation is done (a) from the beginning, (b) after 300 epochs. Although model interpolation and reinitialization happens every 50 epochs, interpolated model accuracy is plotted every epoch. }
        \label{fig:acc_epochs}

\vspace{-0.3cm}
\end{figure}

\textbf{Accuracy across training epochs:} We show the accuracy across training epochs for the individual branches and the combined model in Fig.\ref{fig:acc_epochs} for two cases - (a) performing interpolations from the beginning, and (b) performing interpolations after half the training epochs, as done in DART. It can be noted from (a) that the interpolations in the initial few epochs have poor accuracy since the models are not in a common basin. Further, as seen in initial epochs of (a), when the learning rate is high, SGD training on an interpolated model cannot retain the flat solution due to its implicit bias of moving towards solutions that minimize train loss alone. Whereas, in the later epochs as seen in (b), the improvement obtained after every interpolation is retained. We therefore propose a common training strategy for the initial half of epochs, and split training after that.

\vspace{0.1cm}
\textbf{Ablation experiments:} We note the following observations from the plots in Fig.\ref{fig:num_models} (a-e):
\vspace{0.1cm}
\begin{enumerate}[label=(\alph*),leftmargin=*,noitemsep,nolistsep]
\item \textbf{Effect of Compute:} Using DART, we obtain higher (or similar) performance gains as the number of training epochs increases, whereas the accuracy of ERM+EMA (Mixed) benchmark starts reducing after 300 epochs of training. This can be attributed to the increase in convergence time for learning noisy (or spurious) features due to the intermediate aggregations as shown in Proposition-\ref{prop:Ours_noise}, which prevents overfitting.
\item \textbf{Effect of Interpolation Frequency:} We note that an optimal range of $\lambda$ or the number of epochs between interpolations is 10 - 80, and we set this value to 50. If there is no interpolation for longer epochs, the models drift apart too much, causing a drop in accuracy.
\item \textbf{Effect of Start Epoch:} We note that although the proposed approach works well even if interpolations are done from the beginning, by performing ERM training on mixed augmentations for 300 epochs, we obtain 0.22\% improvement. Moreover, since interpolations do not help in the initial part of training as seen in Fig.\ref{fig:acc_epochs} (a), we propose to start this only in the second half. 
\item \textbf{Effect of Number of branches:} As the number of branches increases, we note an improvement in performance due to higher diversity across branches, leading to more robustness to spurious features and better generalization as shown in Proposition-\ref{prop:SWA}.  
\item \textbf{Effect of Interpolation epochs:} We perform an experiment with 50 epochs of common training followed by a single interpolation. We use a fixed learning rate and plot the accuracy by varying the interpolation epoch. As this value increases, models drift far apart, reducing the accuracy after interpolation. At epoch-500, the accuracy even reaches 0, highlighting the importance of having a low loss barrier between models. 

\end{enumerate}

\vspace{-0.1cm}
 
\section{Conclusion}
In this work, we first show that ERM training using a combination of \emph{diverse} augmentations within a training minibatch can be a strong benchmark for ID generalization, which is outperformed only by ensembling the outputs of individual experts. Motivated by this observation, we present DART - Diversify-Aggregate-Repeat Training, to achieve the benefits of training diverse experts and combining their expertise throughout training. The proposed algorithm first trains several models on different augmentations (or domains) to learn a \emph{diverse} set of features, and further \emph{aggregates} their weights to obtain better generalization. We repeat the steps Diversify-Aggregate several times over training, and show that this makes the optimization trajectory more robust by suppressing the learning of noisy features, while also ensuring a low loss barrier between the individual models to enable their effective aggregation. We justify our approach both theoretically and empirically on several benchmark In-Domain and Domain Generalization datasets, and show that it integrates effectively with several base algorithms as well. We hope our work motivates further research on leveraging the linear mode connectivity of models for better generalization.

\vspace{-0.1cm}
\section{Acknowledgments}
This work was supported by the research grant CRG/2021/005925 from SERB, DST, Govt. of India. Sravanti Addepalli is supported by Google PhD Fellowship.

{\small
\bibliographystyle{ieee_fullname}
\bibliography{references}
}

\newpage
\appendix
\section{Theoretical Results}
\label{sec:theory}
In this section, we present details on the theoretical results discussed in Section-5 \ref{theory_setup}. As noted by Shen \etal \cite{shen2022data}, the weights learned by a patch-wise Convolutional Neural Network are a linear combination of the two types of features (described in Section-\ref{theory_setup} of the main paper) present in the dataset. Let the threshold $K_{cut}$ denote the number of robust features learned by the model. We have,
\begin{equation}
\label{eq:weight_supp}
    w = \sum\limits_{k \leq K_{cut}}v_k + \sum\limits_{k>K_{cut}}y^{(k)}\epsilon^{(k)} 
\end{equation}
\\
On averaging of the weights of $m$ models we get:
\begin{equation}
\label{eq:high_level_weights}
w =  \frac{1}{m}\sum\limits_{j=1}^{m}\left[\sum\limits_{k=1}^{K_{{cut}_j}}v_{k_{j}} + \sum\limits_{k>K_{{cut}_j}}y^{(k)}_j\epsilon^{(k)}_j\right]
\end{equation}

We now analyze the convergence of this weight averaged neural network shown in Eq.\ref{eq:high_level_weights}. Let $L$ represent the logistic loss of the model, $F$ denote the function learned by the neural network, and $w_c$ denote its weights across $C$ channels indexed using $c$. Further, let $y^{(i)}$ represent the ground truth label of sample $x_i ~~ \forall~~ i \in [1,n]$, where $n$ denotes the number of samples in the train set. The weights ${w_1, w_2, .., w_C}$ are initialized as $w_c\sim \mathcal{N}\left(0,\sigma_{0}^2I_d\right) ~ \forall ~c \in C$. We assume that the weights learned by the model at any time stamp $t$ are a linear combination of the linear functions $f$, $g$ and $h$ corresponding to feature patches, noisy patches and model initialization respectively, as shown below:
\begin{equation}
    w_{c}^{t} = f(v_1, v_2, \dots v_K) + g(\epsilon^{(1)}, \epsilon^{(2)}, \dots \epsilon^{(n)}) + h(\epsilon')
\end{equation}
where $\epsilon'$ is the random noise sampled for the initialization of the model. 
Since the term $h(\epsilon')$ does not play a role in the convergence of the model, we ignore this term for the purpose of analysis. For simplicity, we assume that $f$ and $g$ represent summations over their respective arguments. Thus, the weights at any time t can be represented as
\begin{equation}
\label{eq:assumtion_state_model}
 w_{c}^{t} =\sum\limits_{l=1}^{K_{cut}^{t}}\alpha^{t}_lv_l + \sum\limits_{l>K_{cut}^{t}}y^{(l)}\epsilon^{(l)}
\end{equation}
where $K_{cut}^{t}$ and $\alpha^{t}_l$ are a functions of time t. At convergence, $\alpha_i=1~~\forall i \in [1,K_{cut}]$ and $\alpha_i=0$ otherwise.

We now analyze the learning dynamics while training the model. Owing to the gradient descent based updates of model weights over time, the derivative of overall loss $L$ w.r.t. the weights of a given channel $w_c$ can be written as,
\\
\begin{multline}
\label{eq:deriv}
    \frac{d}{dt}w_c = -\frac{d}{d{w_c}}L \\ = -\frac{1}{n}\sum\limits_{i=1}^n y^{(i)} L^{'}(y^{(i)}, F(w, x^{(i)})) \nabla_{w_c}F(w,x^{(i)})
    \end{multline}
\\
Since $L$ is a logistic loss, we have $-L^{'}(o(1)) = 0.5 + o(1)$, where $o(1)$ represents terms independent of the variable $w$. As discussed in Section-\ref{theory_setup} of the main paper, the function learned by the neural network is given by $F(w,x)=\sum\limits_{c=1}^C\sum\limits_{p=1}^2\phi(w_c, x_p)$, where $\phi$ is the activation function defined as follows \cite{shen2022data}:

\begin{itemize}  
\item for $|z|\leq1$; $\phi(z)=sign(z)\frac{1}{q}|z|^q$
\item  for $z\geq1$; $\phi(z)=z - \frac{q-1}{q}$
\item  for $z\leq-1$; $\phi(z)=z + \frac{q-1}{q}$
\end{itemize}

\noindent Based on this, Eq.\ref{eq:deriv} can be written as

\begin{equation}
\frac{d}{dt}w_c \approx \frac{1+o(1)}{2n} \sum\limits_{i=1}^n \sum\limits_{p=1}^2 \phi^{'}(|w_cx_{p}^{(i)}|)y^{(i)}x_{p}^{(i)} 
\end{equation}
\\
Considering the two types of patches present in the image (feature and noisy patch), we have:
\begin{multline}
\label{eq:main}
\frac{d}{dt}w_c \approx \frac{1+o(1)}{2n} \sum\limits_{i=1}^n  \phi^{'}(|w_cv_{d^{(i)}}|)v_{d^{(i)}} \\ + \frac{1+o(1)}{2n} \sum\limits_{i=1}^n  \phi^{'}(|w_c\epsilon^{(i)}|)y^{(i)}\epsilon^{(i)} 
\end{multline}

 where $v_{d^{(i)}}$ represents the feature patch in the image $x^{(i)}$, $\frac{1+o(1)}{2n} \sum\limits_{i=1}^n  \phi^{'}(|w_cv_{d^{(i)}}|)v_{d^{(i)}}$ represents the gradients on feature patches, and $\frac{1+o(1)}{2n} \sum\limits_{i=1}^n  \phi^{'}(|w_c\epsilon^{(i)}|)y^{(i)}\epsilon^{(i)}$ represents the gradients on noisy patches of the image.

To improve the clarity of the proofs, we restate and proof lemma-1 of \cite{shen2022data} in the following two lemmas presented below: \\
\noindent\textbf{Lemma 1} \textit{
\label{lemma:composite_gaussian}
Let $X \sim N(0,\sigma_{x}^2I_{m})$ and $Y \sim N(0,\sigma_{y}^2I_{m})$ be $m$ dimensional gaussian random variables, then $X^{T}Y = O(\sqrt{m}\sigma_{x}\sigma_{y})$}
\begin{proof}
Given any two random variables $x \sim N(\mu_{1}, \sigma_{1}^2)$ and $y \sim N(\mu_{2}, \sigma_{2}^2)$
\begin{multline}
    Var(xy) = E[x^2y^2] - E[(xy)]^{2} = \\ Var(x)Var(y) + Var(x)E(y)^2 + Var(y)E(x)^2 \\
    = \sigma_{1}^{2}\sigma_{2}^{2} + \sigma_{1}^2\mu_{2}^2 + \sigma_{2}^2\mu_{1}^2
\end{multline}
For $\mu_1=\mu_2=0$, we get 
\begin{equation}
\label{eq:product_gauss}
Var(xy) = Var(x)Var(y)
\end{equation}
Let $X \sim N(0,\sigma_{x}^2I_{m})$ and $Y \sim N\left(0,\sigma_{y}^2I_{m}\right)$ be $m$ dimensional gaussian random variables, \ie, $X=[x_0, x_1,  x_2, ..., x_{m-1}]$ and $Y=[y_0, y_1, y_2, ..., y_{m-1}]$, where $x_i  \sim N\left(0, \sigma_{x}^2\right)$ and $y_i  \sim N\left(0, \sigma_{y}^2\right)~ \forall {i\in\{0,1,2, ...,m-1\}}$. Calculating $Var(X^TY)$,
\begin{equation}
Var(X^TY) = E\left[\left(X^TY\right)^2\right] = E\left[\left(\sum\limits_{i=0}^{m-1}x_iy_i\right)^2\right]  
\end{equation}
Since each $x_i$ and $y_i$ are sampled \textit{i.i.d} from a Gaussian with a fixed mean and variance, therefore the product $x_iy_i$ is also an \textit{i.i.d} random variable with a distribution of the difference of two chi-squared distributions. The sum of such $k$ chi-squared random variables with mean $\mu$ and variance $\sigma^2$ results in a chi-squared distribution with mean $k\mu$ and variance $k\sigma^2$. Given this, let $z=X^TY$. Therefore, by Eq.\ref{eq:product_gauss}, $z$ has a zero mean and a variance of $m\sigma_{x}^2\sigma_{y}^2$. Thus, we have
\begin{equation}
    Var(z) = E(z^2) = m\sigma_{x}^2\sigma_{y}^2
\end{equation}
Using Chebyshev's inequality, we have 
\begin{equation}
    P(|z|\ge k\sqrt{m}\sigma_x\sigma_y)\le \frac{1}{k^2}
\end{equation}
where $k$ is some constant. Therefore, we have 
\begin{equation}
\label{eq:chebyshev}
z=O\left(\sqrt{m}\sigma_x\sigma_y\right)
\end{equation}
Further, by central limit theorem, we have the distribution of $z=X^TY = \sum\limits_{i=0}^{m-1}x_iy_i$ to be approximately Gaussian. Therefore, even for a small value of $k$, we have a high confidence interval for bounding $|z|$.
\end{proof}

\noindent\textbf{Lemma 2} \textit{
\label{lemma:composite_gaussian2}
Let $V$ be a standard basis vector and $Y \sim N(0,\sigma_{y}^2I_{m})$ be $N$ dimensional gaussian random variable, then $V^{T}Y = O(\sigma_{y})$}
\begin{proof}
Let $Y \sim N(0,\sigma_{y}^2I_{m})$ be $m$-dimensional gaussian random variable, \ie, $Y=[y_0,y_1, y_2, ..., y_{m-1}]$ where each $y_i  \sim N(0, \sigma_{y}^2)~ \forall {i\in\{0,1,2,...,m-1\}}$. Let $V=[v_0,v_1, v_2, ..., v_{m-1}]$ and $z=V^{T}Y$. Since $V$ is a standard basis vector, we have
\begin{multline}
    Var(z) = E\left[\left(V^{T}Y\right)^2\right] = E\left[\left(\sum\limits_{i=0}^{m-1}v_iy_i\right)^2\right] = \\ E\left[\left(y_k\right)^2\right] = Var(y_k) = \sigma_y^2
\end{multline}
where $k$ is some index for which $v_k=1$ and $v_j=0 ~~ \forall j\neq k$.
Using Chebyshev's inequality, we have 
\begin{equation}
    P(|z|\ge k\sigma_y)\le \frac{1}{k^2}
\end{equation}
where $k$ is some constant. Therefore we have 
\begin{equation}
z=O(\sigma_y)
\end{equation}
\end{proof}
Based on the above lemmas, considering the weights $w_c \sim (0,\sigma_{0}^2I_d)$, we have
\begin{equation}
\label{eq:w_v}
  |w_cv_{k}|  =  O({\sigma_0})
\end{equation}
\begin{equation}
\label{eq:w_eps}
  |w_c\epsilon^{(i)}|  =  O({\sigma\sigma_0})
\end{equation}
\begin{equation}
\label{eq:eps_eps}
  |\epsilon^{(j)}\epsilon^{(i)}|  =  O\left(\frac{\sigma^2}{\sqrt{d}}\right)
\end{equation}
\begin{equation}
\label{eq:eps_v}
  |\epsilon^{(i)}v_k|  = O\left(\frac{\sigma}{\sqrt{d}}\right)
\end{equation}

\subsection{Convergence time for feature patches} 
\noindent \textbf{Data Augmentations:} As defined by Shen \etal \cite{shen2022data}, an augmentation $T_k$ can be defined as follows:
    \begin{multline}
    \label{eq:aug_transform_supp}
\forall ~k'\in[1, K],~~\mathcal{T}_k(v_{k^{'}}) = v_{((k^{'} + k - 1) ~mod~K)+1}
    \end{multline}
    Assuming that $K$ unique augmentation strategies are used (where $K$ denotes the number of robust patches in the dataset), augmented data is defined as follows:
    \begin{multline}
    \label{eq:aug_union_supp}
    D_{train}^{(aug)} = D_{train}~\cup~ \mathcal{T}_1(D_{train}) .. \cup~ \mathcal{T}_{K-1}(D_{train})
    \end{multline}
    where $D_{train}$ is the training dataset.
This ensures that each feature patch $v_i$ appears 
$n$ times in the dataset, thus making the distribution of all the feature patches uniform.
In the proposed method, we consider that $m$ models are being independently trained after which their weights are averaged as shown below: 
\vspace{-0.2cm}
\begin{equation}
\label{eq:weights_watig_supp}
w =  \frac{1}{m}\sum\limits_{j=1}^{m}\sum\limits_{k=1}^{K_{{cut}_j}}v_{k_{j}} + \frac{1}{m}\sum\limits_{j=1}^{m}\sum\limits_{k>K_{{cut}_j}}y^{(k)}_j\epsilon^{(k)}_j
\end{equation}
Each branch is trained on the dataset $D_{train}^{(k)}$ defined as:
    \begin{equation}
    \label{eq:aug_def_supp}
    D_{train}^{(k)} = \mathcal{T}_k(D_{train}), ~~ k\in[1,2,...,m]
    \end{equation}

\noindent\textbf{Proposition 1} \textit{The convergence time for learning any feature patch $v_{i} ~~ \forall i\in[1,K]$ in at least one channel $c \in C$ of the weight averaged model  $f_\theta$ using the augmentations defined in Eq.\ref{eq:aug_def_supp}, is given by $O\left(\frac{K}{\sigma_{0}^{q-2}}\right)$, if $\frac{\sigma^q}{\sqrt{d}} \ll \frac{1}{K}$, $m=K$.}
\begin{proof}
We first compute the convergence time without weight-averaging, as shown by Shen \etal \cite{shen2022data}. The dot product between $\frac{dw_c}{dt}$ (from Eq.\ref{eq:main}) and any given feature  $v_k$ is given by:
\begin{multline}
\label{eq:v_k_multi}
    \frac{d}{dt}w_c\! \cdot \!v_k \approx \frac{1+o(1)}{2}\rho_k  \phi^{'}(|w_cv_{k}|) \\ + \frac{1+o(1)}{2n} \sum\limits_{i=1}^n  \phi^{'}(|w_c\epsilon^{(i)}|)y^{(i)}\epsilon^{(i)}v_k 
\end{multline}
where, $\rho_k$ represents the fraction of $v_k$ in the dataset.
At initialization, we have $w_c\! \sim \!(0,\sigma_{0}^2I_d)$. Therefore, using conditions at initialization in Eq.\ref{eq:w_v}, \ref{eq:w_eps} and \ref{eq:eps_v} along with the definition of the activation function defined for the case $|w_cv_{d^{(i)}}|\!<\!1$ and $|w_c\epsilon^{(i)}|\!<\!1$, we arrive at the following convergence time for the feature and the noisy patch, respectively:
\begin{equation}
\label{eq:patch_bound}
   \frac{1+o(1)}{2n}  \phi^{'}\!\left(|w_cv_{k}|\right)\! =\!  O\left({\rho_k\sigma_0^{q-1}}\right)
\end{equation}
\begin{equation}
\label{eq:noise_bound}
   \frac{1+o(1)}{2n} \sum\limits_{i=1}^n  \phi^{'}(|w_c\epsilon^{(i)}|)y^{(i)}\epsilon^{(i)}v_k \! =\!  O\!\left(\frac{\sigma_0^{q-1}\sigma^{q}}{\sqrt{d}}\right)
\end{equation}
A closer look at the above two equations reveal that if $\frac{\sigma^q}{\sqrt{d}} \ll \frac{1}{K}$, the noisy patch term in Eq.\ref{eq:v_k_multi} (the second term) can be ignored in comparison to the feature patch term (the first term). This gives:
\begin{equation}
\label{eq:good_patch}
\frac{d}{dt}w_c\! \cdot \!v_{k} \approx \frac{1+o(1)}{2} \rho_k\phi^{'}(|w_cv_{k}|) 
\end{equation}
\noindent  Let us denote the term $w_c\! \cdot \!v_k$ at any time step $t$ using a generic function  $g \equiv g(w_c,v_k,t)$. Using the definition of the activation function $\phi$, and assuming that $|w_cv_k|\!<\!1$, we get
\begin{equation}
\frac{dg}{dt}\! =\! \frac{1+o(1)}{2}{\rho_k} g^{q-1}
\end{equation}
On integrating, we get the following:
\begin{multline}
\label{eq:integrate}
\frac{(1\!+\!o(1))\rho_k}{2}(2-q)t + g(t\!=\!0)^{2-q} = g(t\!=\!t)^{2-q}
\end{multline}
\begin{equation}
\label{eq:conv_time_feat}
t =  O\left(\frac{1}{\rho_k\sigma_0^{q-2}}\right)
\end{equation}
We now compute the convergence of the case where $m$ models are averaged. We denote the averaged weights of a given channel $c$ by $w_{c}^{avg}$. By substituting for $w_c$ from Eq.\ref{eq:assumtion_state_model}, we get
\begin{multline}
\label{eq:good_patch_wa}
-\frac{1}{m}\sum\limits_{j=1}^m{\left(\frac{dL}{dw_{c}}\right)}_j v_k=\frac{dw_{c}^{avg}}{dt}v_k \\= \frac{1}{m}\sum\limits_{j=1}^{m}\frac{d}{dt}\left(\sum\limits_{l=1}^{K}\alpha^{t}_{lj}v_l +\! \sum\limits_{l>K_{cut}^{t}}y^{(l)}_j\epsilon^{(l)}_j\right)v_k
\end{multline}

Using $|\epsilon^{(i)}v_k|\! =\! O\left(\frac{\sigma}{\sqrt{d}}\right)$ from Eq.\ref{eq:eps_v} gives us $\sum\limits_{l>K_{cut}^{t}}y^{(l)}\epsilon^{(l)}v_{k}=O\left(\frac{\sigma}{\sqrt{d}}\right)$, whereas $\sum\limits_{l=1}^{K}\alpha^{t}_{lj}v_l\!=\!O(1)$. Since $d$ represents the number of parameters, we can say $d\!\gg\!\sigma$. Further, since $\epsilon^{(l)}$ are \textit{i.i.d} random variables, therefore, the value of the noise component $\sum\limits_{l>K_{cut}^{t}}y^{(l)}\epsilon^{(l)}v_{k}$ is expected to further decrease upon averaging over $m$ models. Thus, ignoring it w.r.t. to the feature term  $\sum\limits_{l=1}^{K}\alpha^{t}_{lj}v_l$, we get
\begin{equation}
\frac{dw_{c}^{avg}}{dt}v_k  \approx \frac{1}{m}\frac{d}{dt}\left(\sum\limits_{j=1}^{m}\alpha^{t}_{kj}\right)
\end{equation} 
A similar analysis for a single model that is not weight-averaged gives
\begin{equation}
\label{eq:corollary}
\frac{dw_{c}}{dt}v_k  \approx  \frac{d\alpha^{t}_{k}}{dt} = \frac{dw_{c}^{avg}}{dt}v_k\frac{d\left(m\alpha^{t}_k\right)}{d\left(\sum\limits_{j=1}^{m}\alpha^{t}_{kj}\right)}
\end{equation} 
As discussed in Section-\ref{theory_setup} of the main paper, we set $m=K$. Further, since the most frequent patches are learned faster, we assume that the relative rate of change in $\alpha_{kj}$ will depend on the relative frequency of individual patch features. Therefore, $\frac{d\alpha^{t}_k/dt}{d\left(\sum\limits_{j=1}^{m}\alpha^{t}_{kj}\right)/dt} = \frac{d\alpha^{t}_k/dt}{d\left(\sum\limits_{j=1}^{K}\alpha^{t}_{kj}\right)/dt} = \rho_k$. Thus we get,
\begin{equation}
\label{eq:grad_w}
    \frac{dw_{c}^{avg}}{dt}v_k = \frac{1}{\rho_k K}\frac{dw_{c}}{dt}v_k
\end{equation}
In Eq.\ref{eq:grad_w}, we have the rate of change of $w_{c}^{avg} \!=\! \frac{1}{\rho_k K}$ times the rate of change of $w_{c}$. Therefore the time for convergence for $w_{c}^{avg}$ will be ${\rho_k K}$ times the time for convergence for $w_{c}$, which gives
\begin{equation}
t =  O\left(\frac{K}{\sigma_0^{q-2}}\right)
\end{equation}
\end{proof}

\noindent\textbf{Corollary 1.1} \textit{
\label{prop-corollary}
The convergence time for learning any feature patch $v_{i} ~~ \forall i\in[1,K]$ in at least one channel $c \in C$ of the weight averaged model  $f_\theta$ using the augmentations defined in Eq.\ref{eq:aug_def_supp}, is given by $O\left(\frac{m\rho_k^{'}}{\rho_k\sigma_{0}^{q-2}}\right)$, if $\frac{\sigma^q}{\sqrt{d}} \ll \frac{1}{K}$.}
Here $\rho_k$ is the ratio between the frequency of the feature patch $k$ in the dataset and the sum of the frequencies of all feature patches in the dataset. $\rho_k^{'}$ is the ratio between the frequency of the feature patch $k$ in the dataset and the sum of the frequencies of some $m$ feature patches $[v_{(k) ~mod~K+1}, v_{(k + 1) ~mod~K+1}, ..., v_{(m + k - 1) ~mod~K+1}]$

\begin{proof}
    Since the most frequent patches are learned faster, we assume that the relative rate of change in $\alpha_{kj}$ will depend on the relative frequency of individual patch features. Therefore, 
    \begin{equation}
    \frac{d\alpha^{t}_k/dt}{d\left(\sum\limits_{j=1}^{m}\alpha^{t}_{kj}\right)/dt} = \frac{(\alpha^{t}_k)/dt}{\left(\sum\limits_{j=1}^{m}\alpha^{t}_{kj}\right)/dt} = \rho_k^{'}
    \end{equation}
    
Thus substituting in Eq.\ref{eq:corollary}, we get
\begin{equation}
\label{eq:grad_w_modified}
    \frac{dw_{c}^{avg}}{dt}v_k = \frac{1}{\rho_k^{'}m}\frac{dw_{c}}{dt}v_k
\end{equation}

In Eq.\ref{eq:grad_w_modified}, we have the rate of change of $w_{c}^{avg} = \frac{1}{\rho_k^{'} m}$ times the rate of change of $w_{c}$. Therefore, the time for convergence for $w_{c}^{avg}$ will be ${\rho_k^{'} m}$ times the time for convergence for $w_{c}$, which gives
\begin{equation}
\label{eq:bound_corollary}
t =  O\left(\frac{m\rho_k^{'}}{\rho_k\sigma_{0}^{q-2}}\right)
\end{equation}
\end{proof}

The convergence time from corollary-\ref{prop-corollary} (denoted as $t$) can be written as
\begin{equation}
\label{eq:time_new}
    t = O\left(\frac{m\sum\limits_{j=1}^{K}\alpha_{kj}}{\sum\limits_{j=1}^{m}\alpha_{kj}\sigma_{0}^{q-2}}\right)
\end{equation}
The convergence time from Eq.\ref{eq:conv_time_feat} (denoted as $t^{'}$) can be written as
\begin{equation}
\label{eq:time_old}
    t^{'} = O\left(\frac{\sum\limits_{j=1}^{K}\alpha_{kj}}{\alpha_{k}\sigma_{0}^{q-2}}\right)
\end{equation}
 For hard to learn feature patches (feature patches with low $\alpha_k$), upon comparing Eq.\ref{eq:time_new} and Eq.\ref{eq:time_old}, we observe that the convergence time will be higher in Eq.\ref{eq:time_old}. Since a summation over some $m$ feature patches is appearing in Eq.\ref{eq:time_new}, therefore its convergence time has a lower impact on the frequency of an individual feature patch. This helps in enhanced learning of hard features, thereby improving generalization.

\subsection{Convergence time of noisy patches}
We consider the dot product between any noisy patch $\epsilon^{k}$ and Eq.\ref{eq:main}:

\begin{multline}
\label{eq:noisy_patch}
\frac{d}{dt}w_c \epsilon^{(k)}=\frac{1+o(1)}{2n} \sum\limits_{i=1}^n  \phi^{'}(|w_cv_{d^{(i)}}|)v_{d^{(i)}}\epsilon^{(k)} + \\ \frac{1+o(1)}{2n} \sum\limits_{i=1}^n  \phi^{'}(|w_c\epsilon^{(i)}|)y^{(i)}\epsilon^{(i)}\epsilon^{(k)}
\end{multline}
On simplifying we get,
\begin{multline}
\label{eq:noise_eq}
\frac{d}{dt}w_c \epsilon^{(k)}=\frac{1+o(1)}{2n} \sum\limits_{i=1}^n  \phi^{'}(|w_cv_{d^{(i)}}|)v_{d^{(i)}}\epsilon^{(k)} + \\ \frac{1+o(1)}{2n} \phi^{'}(|w_c\epsilon^{(k)}|)y^{(k)}||\epsilon^{(k)}||^2 \\ + \frac{1+o(1)}{2n} \sum\limits_{i=1;i\neq k}^n  \phi^{'}(|w_c\epsilon^{(i)}|)y^{(i)}\epsilon^{(i)}\epsilon^{(k)}
\end{multline}

In Eq.\ref{eq:noise_eq} we can ignore $\frac{1+o(1)}{2n} \sum\limits_{i=1}^n  \phi^{'}(|w_cv_{d^{(i)}}|)v_{d^{(i)}}\epsilon^{(k)}\\ + \frac{1+o(1)}{2n} \sum\limits_{i=1;i\neq k}^n  \phi^{'}(|w_c\epsilon^{(i)}|)y^{(i)}\epsilon^{(i)}\epsilon^{(k)}$ as compared to 
$\frac{1+o(1)}{2n} \phi^{'}(|w_c\epsilon^{(k)}|)y^{(k)}||\epsilon^{(k)}||^2$, if their values are of different orders at initialization. Since, at initialization, $w_c \sim (0,\sigma_{0}^2I_d)$, using conditions in Eq.\ref{eq:w_v}-\ref{eq:eps_v} and the definition of the activation function defined for the case of $|w_cv_{d^{(i)}}|<1$ and $|w_c\epsilon^{(i)}|<1 \forall i \in [1,n]$, we get
\begin{equation}
   \frac{1+o(1)}{2n} \sum\limits_{i=1}^n  \phi^{'}(|w_cv_{d^{(i)}}|)v_{d^{(i)}}\epsilon^{(k)} =  O\left({\frac{\sigma_0^{q-1}\sigma}{\sqrt{d}}}\right)
\end{equation}
\begin{equation}
   \frac{1+o(1)}{2n}\! \sum\limits_{i=1;i\neq k}^n  \phi^{'}(|w_c\epsilon^{(i)}|)y^{(i)}\epsilon^{(i)}\epsilon^{(k)} =  O\left(\frac{\sigma_0^{q-1}\
   \sigma^{q+1}}{\sqrt{d}}\right)
\end{equation}
Therefore,
\begin{multline}
\label{eq:noise_bound2}
   \frac{1+o(1)}{2n} \sum\limits_{i=1}^n  \phi^{'}(|w_cv_{d^{(i)}}|)v_{d^{(i)}}\epsilon^{(k)} 
 + \\  \frac{1+o(1)}{2n} \sum\limits_{i=1;i\neq k}^n  \phi^{'}(|w_c\epsilon^{(i)}|)y^{(i)}\epsilon^{(i)}\epsilon^{(k)} = \\  O\left(\frac{\sigma_0^{q-1}\
   \sigma^{q+1}}{\sqrt{d}}\right) + O\left({\frac{\sigma_0^{q-1}\sigma}{\sqrt{d}}}\right)
\end{multline}
\begin{multline}
\label{eq:patch_bound2}
\frac{1+o(1)}{2n} \phi^{'}(|w_c\epsilon^{(k)}|)y^{(k)}||\epsilon^{(k)}||^2 = \\ \frac{1+o(1)}{2n} \sigma^2 \phi^{'}(|w_c\epsilon^{(k)}|)y^{(k)} = O\left(\frac{\sigma^{q+1}\sigma_0^{q-1}}{n}\right)
\end{multline}
Comparing Eq.\ref{eq:patch_bound2} and Eq.\ref{eq:noise_bound2}, we get if $d\! \gg \!n^2$, we can ignore the term $\frac{1+o(1)}{2n} \sum\limits_{i=1}^n  \phi^{'}(|w_cv_{d^{(i)}}|)v_{d^{(i)}}\epsilon^{(k)} 
 +  \frac{1+o(1)}{2n} \sum\limits_{i=1;i\neq k}^n  \phi^{'}(|w_c\epsilon^{(i)}|)y^{(i)}\epsilon^{(i)}\epsilon^{(k)}$ as compared to $\frac{1+o(1)}{2n} \phi^{'}(|w_c\epsilon^{(k)}|)y^{(k)}||\epsilon^{(k)}||^2$ Thus, we get the following:

Using the activation defined earlier, and considering the value of $w_c\epsilon^{(k)}$ at time stamp $t$ given by $g(w_c,\epsilon^{(k)},t)$, where $|g(w_c,\epsilon^{(k)}, t)|<1$, we get
\begin{equation}
\label{eq:diffeq}
\frac{d(g(w_c,\epsilon^{(k)}, t))}{dt} = \frac{1+o(1)}{2n}{\sigma^2} g(w_c,\epsilon^{(k)}, t)^{q-1}
\end{equation}
Similar to the analysis presented in Eq.\ref{eq:integrate}, on integrating the above equation, we get 
\begin{multline}
\label{eq:convtime}
\frac{1+o(1)}{2n}(2-q)t\sigma^2 + g(w_c,\epsilon^{(k)},t=0)^{2-q}  = \\ g(w_c,\epsilon^{(k)},t=t)^{2-q}
\end{multline}
Using Eq.\ref{eq:w_eps}, at $t=0$,
\begin{equation}
\label{eq:init_value}
g(w_c,\epsilon^{(k)},t=0)^{2-q}=\sigma_0^{2-q}\sigma^{2-q}
\end{equation}
where $\sigma_0$ is the standard deviation of the zero-mean Gaussian distribution that is used for initializing the weights of the model, and $\frac{\sigma}{\sqrt{d}}$ is the standard deviation of the noise present in noisy patches. Thus, we get
\begin{multline}
\frac{1+o(1)}{2n}(2-q)t\sigma^2 + \sigma_0^{2-q}\sigma^{2-q}  = \\  g(w_c,\epsilon^{(k)},t=t)^{2-q}
\end{multline}

At the time of convergence, the term $g(w_c,\epsilon^{(k)},t=t)^{2-q}$ will become $o(1)$. Therefore, $\frac{1+o(1)}{2n}(2-q)t\sigma^2 + \sigma_0^{2-q}\sigma^{2-q}$ should be constant. Equating the L.H.S. of the above equation to 0, the convergence time to learn $\epsilon^{(k)}$ by at least one channel $c\in C$ is given by:
\begin{equation}
\label{eq:results_noisy_patches}
t =  O\left(\frac{n}{\sigma_0^{q-2}\sigma^q}\right)
\end{equation}

\noindent\textbf{Proposition 2} \textit{
\label{prop:SWA_supp}
If the noise patches learned by each $f_\theta^k$ are $i.i.d.$ Gaussian random variables $\sim \mathcal{N}\left(0, \frac{\sigma^2}{d}I_d\right)$ then with high probability, convergence time of learning a noisy patch $\epsilon^{(j)}$ in at least one channel $c \in [1,C]$ of the weight averaged model $f_\theta$ is given by $O\left(\frac{nm}{\sigma_{0}^{q-2}\sigma^q}\right)$, if $d \gg n^2$.}
\begin{proof}
By averaging the weights of $m$ models in Eq.\ref{eq:noisy_patch}, we get
\begin{multline}
- \frac{1}{m}\sum\limits_{j=1}^m{\left(\frac{dL}{dw_{c}}\right)}_j\epsilon^{(k)}= \frac{dw_{avg}}{dt}\epsilon^{(k)} = \\ \frac{1}{m}\sum\limits_{j=1}^m\bigg[\frac{1+o(1)}{2n} \sum\limits_{i=1}^n  \phi^{'}(|w_{c_j}v_{d^{(i)}}|)v_{d^{(i)}}\epsilon^{(k)} +\\ \frac{1+o(1)}{2n} \sum\limits_{i=1}^n  \phi^{'}(|w_{c_j}\epsilon^{(i)}|)y^{(i)}\epsilon^{(i)}\epsilon^{(k)}\bigg]
\end{multline}

\begin{multline}
=\frac{1}{m}\sum\limits_{j=1}^m\bigg[\frac{1+o(1)}{2n} \sum\limits_{i=1}^n  \phi^{'}(|w_{c_j}v_{d^{(i)}}|)v_{d^{(i)}}\epsilon^{(k)} +  \\ \frac{1+o(1)}{2n}  \phi^{'}(|w_{c_j}\epsilon^{(k)}|)y^{(k)}||\epsilon^{(k)}||^{2}_{2} + \\ \frac{1+o(1)}{2n} \sum\limits_{i=1; i\neq k}^n  \phi^{'}(|w_{c_j}\epsilon^{(i)}|)y^{(i)}\epsilon^{(i)}\epsilon^{(k)}\bigg]
\end{multline}
\begin{equation}
=\sum\limits_{j=1}^m\frac{1}{m}\bigg[ \frac{1+o(1)}{2n}\sigma^2  \phi^{'}(|w_{c_j}\epsilon^{(k)}|)y^{(k)} +\tau \bigg]
\end{equation}
where $\tau$ consists of the remaining terms that are negligible since the noise learned by each model is $i.i.d.$ and $d \gg n^2$.
Using the weights learned by different models as represented in Eq.\ref{eq:assumtion_state_model}, we get,
\begin{multline}
\label{eq:noisy_patch_wa}
\frac{dw_{avg}}{dt}\epsilon^{(k)} \approx \frac{1+o(1)}{2n}\sigma^2y^{(k)}\frac{1}{m}  \sum\limits_{j=1}^m\phi^{'}\big(\big|(\sum\limits_{l=1}^{K}\alpha^{t}_{lj}v_l \\ +  \sum\limits_{l>K_{cut}^{t}}y^{(l)}_{j}\epsilon^{(l)}_{j}) \epsilon^{(k)}\big|\big) 
\end{multline}
Since the noise $\epsilon^{(i)}$ learned by different models is considered as $i.i.d.$, we get
\begin{multline}
\frac{dw_{avg}}{dt}\epsilon^{(k)} = \frac{1+o(1)}{2nm}\bigg(\sigma^2y^{(k)}\phi^{'}\big(|\sum\limits_{l=1}^{K}\alpha^{t}_{lk}v_l\epsilon^{(k)} +\\ \sum\limits_{l>K_{cut}^{t}}y^{(l)}_{k}\sigma^2|\big) + \sum\limits_{i=1; i\neq k}^m\phi^{'}\big(|\sum\limits_{l=1}^{K}\alpha^{t}_{li}v_l\epsilon^{(k)} + \\ \sum\limits_{l>K_{cut}^{t}}y^{(l)}_{i}\epsilon^{(l)}_{i}\epsilon^{(k)}|\big) \bigg)
\end{multline}
Note that from Eq.\ref{eq:eps_eps}, we have $\sum\limits_{l>K_{cut}^{t}}y^{(l)}\epsilon^{(l)}_{i}\epsilon^{(k)}=O\left(\frac{\sigma^2}{\sqrt{d}}\right)$, and from Eq.\ref{eq:eps_v}, we get $\sum\limits_{i=1; i\neq k}^n\phi^{'}\left(\left|\sum\limits_{l=1}^{K}\alpha^{t}_{li}v_l\epsilon^{(k)}\right|\right)=O\left(\left(\frac{\sigma}{\sqrt{d}}\right)^{q-1}\right)$. Whereas $y^{(l)}_{k}\sigma^2=O(1)$. Since it is assumed that $d \gg n^2$, therefore, we can ignore the terms $\sum\limits_{l=1}^{K}\alpha^{t}_{li}v_l\epsilon^{(k)}$ and $\sum\limits_{l>K_{cut}^{t}}y^{(l)}\epsilon^{(l)}_{i}\epsilon^{(k)}$ in comparison to $\sum\limits_{l>K_{cut}^{t}}y^{(l)}_{k}\sigma^2$. Thus, we get
\begin{multline}
\label{eq:weight_model_dynamics}
\frac{dw_{avg}}{dt}\epsilon^{(k)} = \frac{1+o(1)}{2nm}\sigma^2y^{(k)}\phi^{'}(|\sum\limits_{l=1}^{K}\alpha^{t}_{lk}v_l\epsilon^{(k)} + \\ \sum\limits_{l>K_{cut}^{t}}y^{(l)}_{k}\sigma^2|)
\end{multline}

Similarly, we derive the learning dynamics of a single model $w_{c_{k}}$ below:
\begin{multline}\label{eq:single_model_dynamics}
\frac{dw_{c_{k}}}{dt}\epsilon^{(k)} = \frac{1+o(1)}{2n}\sigma^2y^{(k)}\phi^{'}(|\sum\limits_{l=1}^{K}\alpha^{t}_{lk}v_l\epsilon^{(k)} + \\ \sum\limits_{l>K_{cut}^{t}}y^{(l)}_{k}\sigma^2|) 
\end{multline}
From Eq.\ref{eq:weight_model_dynamics} and Eq.\ref{eq:single_model_dynamics}, we get the following relation
\begin{equation}
\label{eq:final_prop2}
\frac{1}{m}\frac{dw_{c_{k}}}{dt}\epsilon^{(k)} = \frac{dw_{avg}}{dt}\epsilon^{(k)}
\end{equation}
In Eq.\ref{eq:final_prop2}, we have the rate of change of $w_{avg}$ equals $\frac{1}{m}$ times the rate of change of $w_{c_{k}}$. Therefore, the time for convergence for $w_{avg}$ will be ${m}$ times the time for convergence for $w_{c_{k}}$, which gives
the convergence time for learning the noisy patch, $\epsilon^{(k)}$ by at least one channel $c \in C$ of the model as
\begin{equation}
t = O\left(\frac{nm}{\sigma_{0}^{q-2}\sigma^q}\right)
\end{equation}
\end{proof}
\noindent\textbf{Proposition 3} 
\label{prop:Ours_noise_supp}
\textit{ If the noise learned by each $f_\theta^k$ are $i.i.d.$ Gaussian random variables $\sim \mathcal{N}\left(0, \frac{\sigma^2}{d}I_d\right)$, and model weight averaging is performed at epoch $T$, the convergence time of learning a noisy patch $\epsilon^{(j)}$ in at least one channel $c \in [1,C]$ of the weight averaged model $f_\theta$ is given by $T + O\left(\frac{nm^{(q-2)}d^{(q-2)/2}}{\sigma^{(2q-2)}}\right)$,
if $d \gg n^2$.}
\begin{proof}
We assume that the model is close to convergence at epoch $T$. Hence, its weights can be assumed to be similar to Eq.\ref{eq:high_level_weights}.  

Further, we assume that the weights are composed of noisy and feature patches as shown in Eq.\ref{eq:high_level_weights}. Since the noisy patches are assumed to be $i.i.d.$, the standard deviation of the weights corresponding to noisy features is given by $\frac{\sigma}{m\sqrt{d}}$. Thus, using the above lemmas, we get
\begin{equation}
g(w_c,\epsilon^{(k)},t=T)^{2-q} = \sigma^{4-2q}m^{q-2}d^{\frac{q-2}{2}}
\end{equation}
On integrating Eq.\ref{eq:diffeq} from time $T$ and substituting the above, we get

\begin{multline}
\frac{1+o(1)}{2n}t(2-q)\sigma^2 + \sigma^{4-2q}m^{q-2}d^{\frac{q-2}{2}} = \\ g(w_c,\epsilon^{(k)},t=t)^{2-q}
\end{multline}
Thus, the convergence time of learning at least one channel $c \in C$ by on using this initialization is given by 
\begin{equation}
t = O\left(\frac{nm^{(q-2)}d^{(q-2)/2}}{\sigma^{(2q-2)}}\right)   
\end{equation}
Further, the total convergence time is given by 
\begin{equation}
T + O\left(\frac{nm^{(q-2)}d^{(q-2)/2}}{\sigma^{(2q-2)}}\right)   
\end{equation}
Since we have considered the weights to be composed of two parts and the model is assumed to be converged with respect to feature patches, therefore, using such an initialization will not impact their learning dynamics.
\end{proof}

\subsection{Impact of intermediate interpolations}

We assume that $T$ in Proposition-3 is negligible w.r.t. $O\left(\frac{nm^{(q-2)}d^{(q-2)/2}}{\sigma^{(2q-2)}}\right)$. We further analyze the ratio of the convergence time from Proposition-3 (denoted as $t$) and Proposition-2 (denoted as $t'$), 
\begin{equation}
\label{eq:lower_bound_analysis}
    \frac{t}{t'} = O\left(\frac{m^{q-3}d^{(q-2)/2}\sigma_0^{q-2}}{\sigma^{q-2}}\right) 
\end{equation}
A lower bound on the above equation will occur when $d=n^2$ and $q=3$. Using this, we get
\begin{equation}
    \frac{t}{t'} = O\left(\frac{n\sigma_0}{\sigma}\right) 
\end{equation}
Thus, the lower bound is of the order $n$ which is greater than 1. Therefore, the convergence time of learning a noisy patch $\epsilon^{(j)}$ in at least one channel $c \in [1,C]$ on performing an intermediate interpolation (Prop.3) is greater than the case where weight-averaging of only final models is performed (Prop.2), by upto $O(n)$.

\begin{figure}
\centering
\hfill{}
\begin{subfigure}[b]{0.2\textwidth}
\centering
\includegraphics[width=\textwidth]{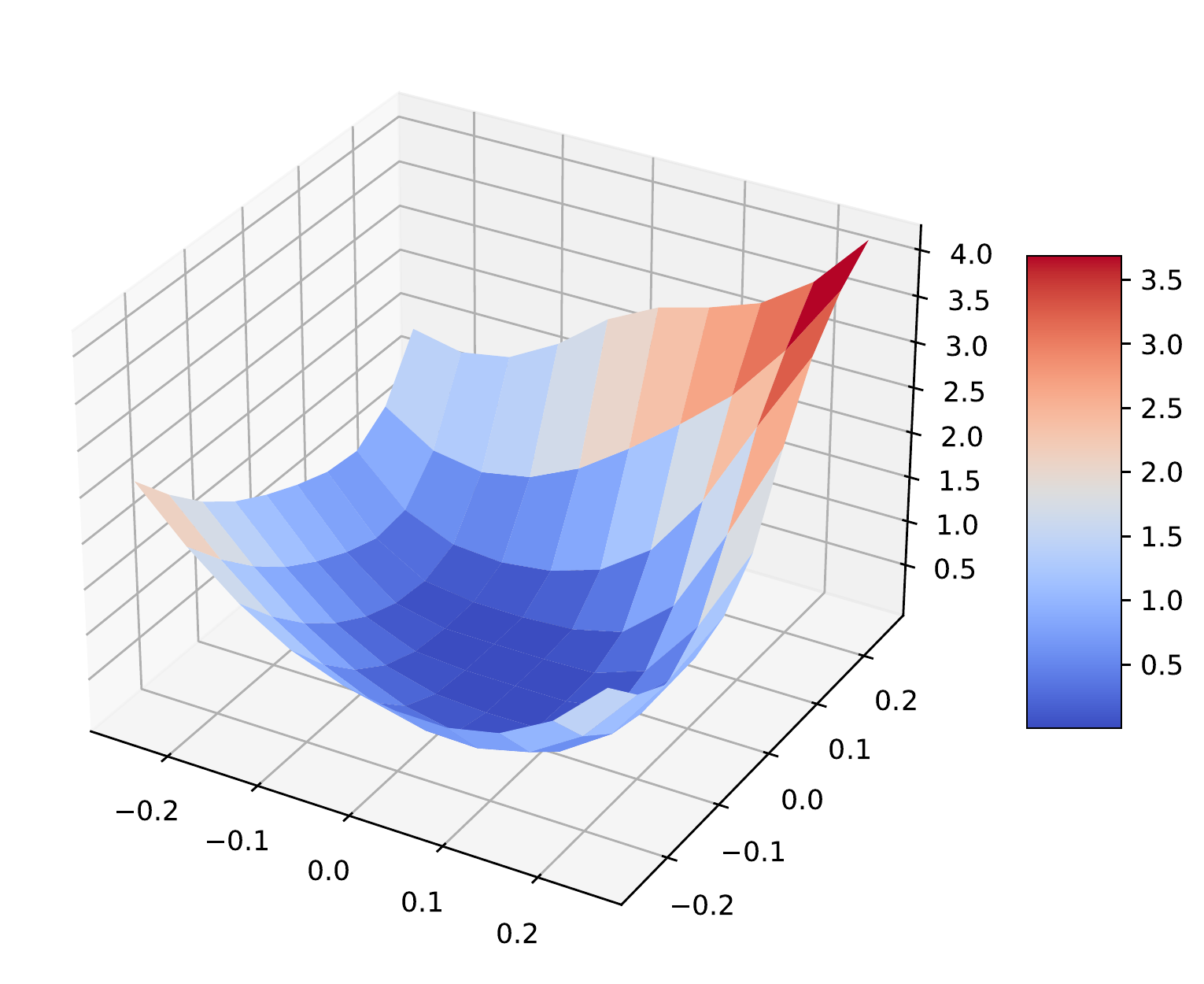}
\caption{ERM+EMA (PCH)}
\label{fig:pc_ema_loss_viz}
\end{subfigure}
\hfill
\begin{subfigure}[b]{0.2\textwidth}
\centering
\includegraphics[width=\textwidth]{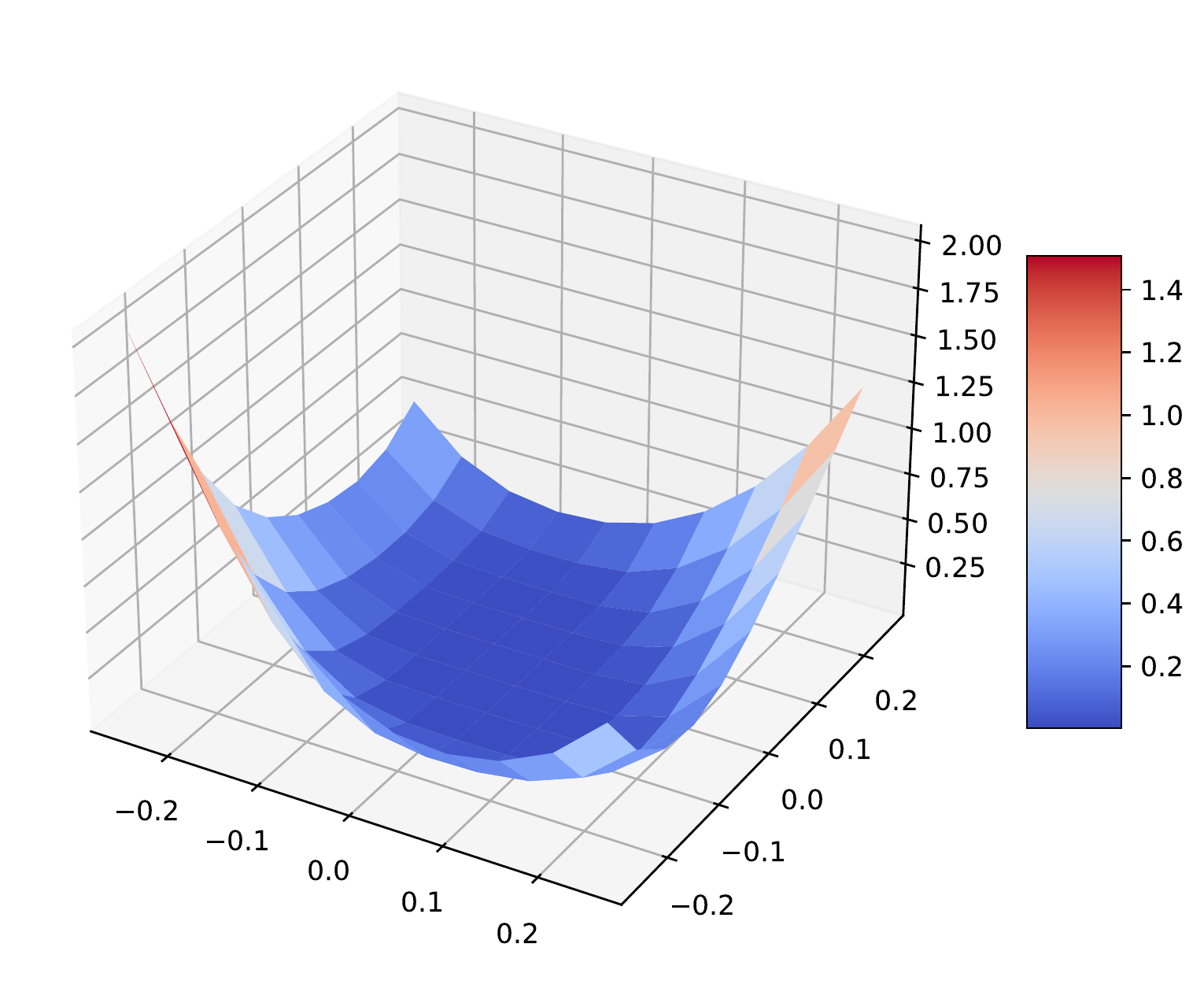}
\caption{DART (Ours, PCH)}
\label{fig:dart_loss_viz}
\end{subfigure}
\hfill{}
\caption{\textbf{Loss landscape visualization}}
\label{fig:loss_landscape_viz}
\end{figure}

\begin{figure}
\centering
\hfill{}
\begin{subfigure}[b]{0.2\textwidth}
\centering
\includegraphics[width=\textwidth]{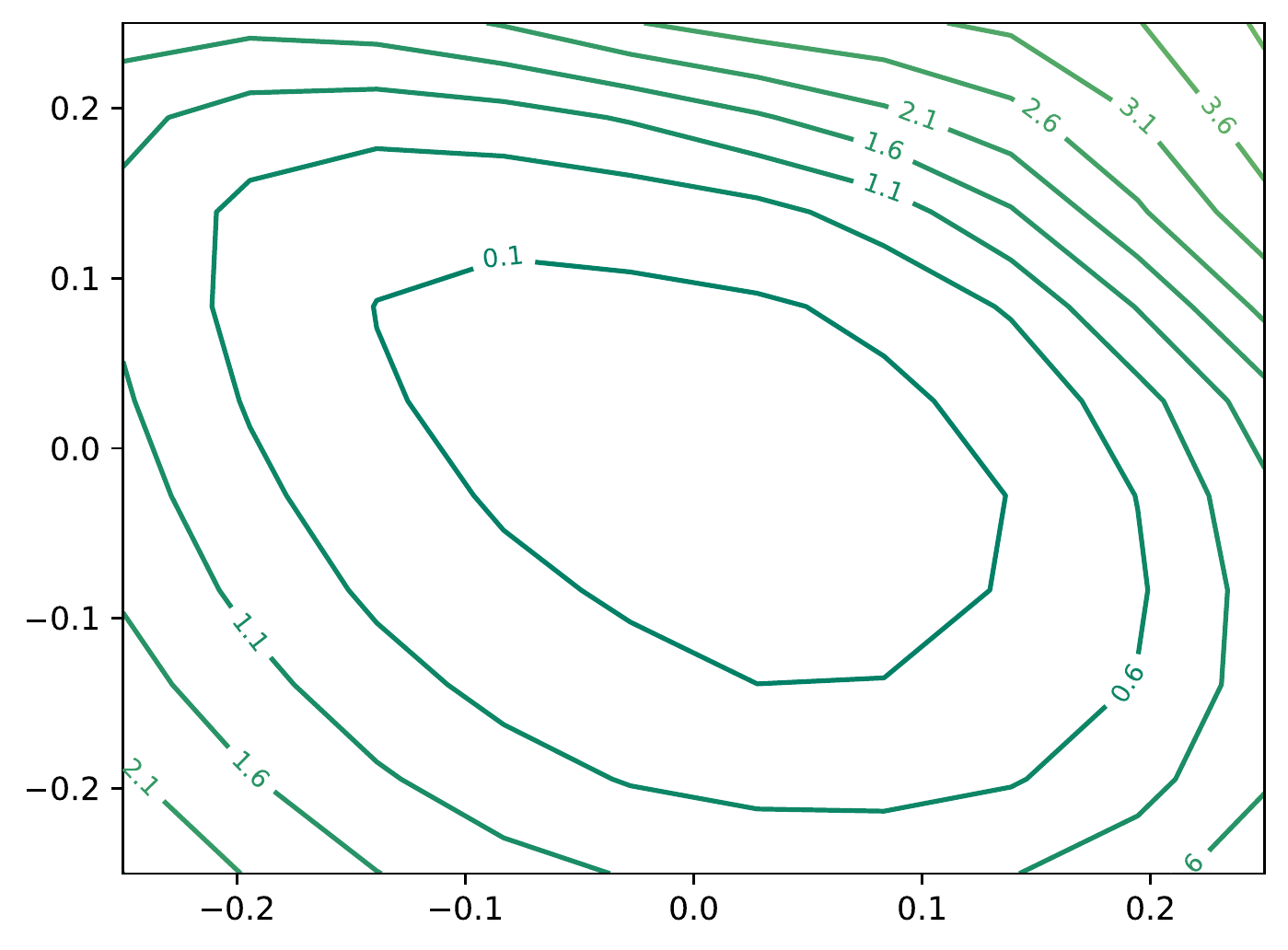}
\caption{ERM+EMA (PCH)} 
\label{fig:pc_ema_contour}
\end{subfigure}
\hfill
\begin{subfigure}[b]{0.2\textwidth}
\centering
\includegraphics[width=\textwidth]{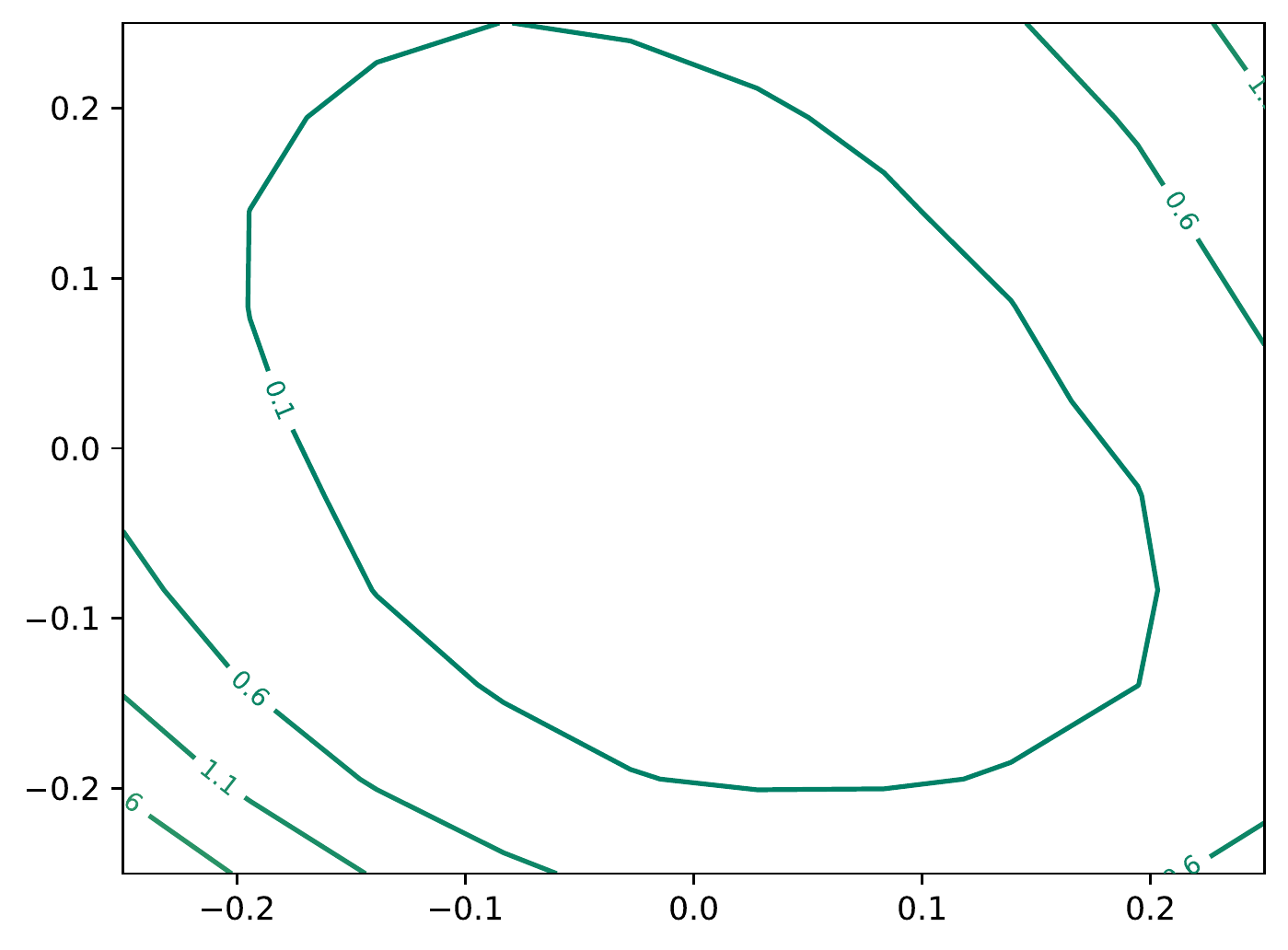}
\caption{DART (Ours, PCH)}
\label{fig:dart_contour}
\end{subfigure}
\hfill{}
\caption{\textbf{Loss Contour visualization}}
\label{fig:loss_contour_viz}
\vspace{-0.3cm}
\end{figure}

\section{Loss surface plots}

We compare the loss surface of the proposed method with ERM training on CIFAR-100 dataset using WRN-28-10 architecture. To exclusively understand the impact of the proposed Diversify-Aggregate-Repeat steps, we present results using the simple augmentations - Pad and Crop followed by Horizontal Flip (PCH) for both ERM and DART. We use exponential moving averaging (EMA) of weights in both the ERM baseline and DART for a fair comparison. 

As shown in Fig.\ref{fig:loss_landscape_viz}, the loss surface of the proposed method DART is flatter when compared to the ERM baseline.  The same is also evident from the level sets of the contour plot in Fig.\ref{fig:loss_contour_viz}. In Table-\ref{table:loss_landscape}, we also use the scale-invariant metrics proposed by Stutz \etal \cite{stutz2021relating} to quantitatively verify that the flatness of loss surface is indeed better using the proposed approach DART. Worst Case Flatness represents the Cross-Entropy loss on perturbing the weights in an $\ell_2$ norm ball of radius $0.25$. Average Flatness represents the Cross-Entropy loss on adding random Gaussian noise with standard deviation $0.25$, and further clamping it so that the added noise remains within the $\ell_2$ norm ball of radius $0.25$. Average Train Loss represents the loss on train set images as shown in Table-\ref{table:loss_landscape}. We achieve lower values when compared to the ERM baseline across all metrics, demonstrating that the proposed method DART has a flatter loss landscape compared to ERM.
\input{tables/loss_landscape}

\section{Additional Results: ID generalization}

\subsection{Model coefficients}

\begin{figure}
\centering
        \includegraphics[width=1\linewidth]{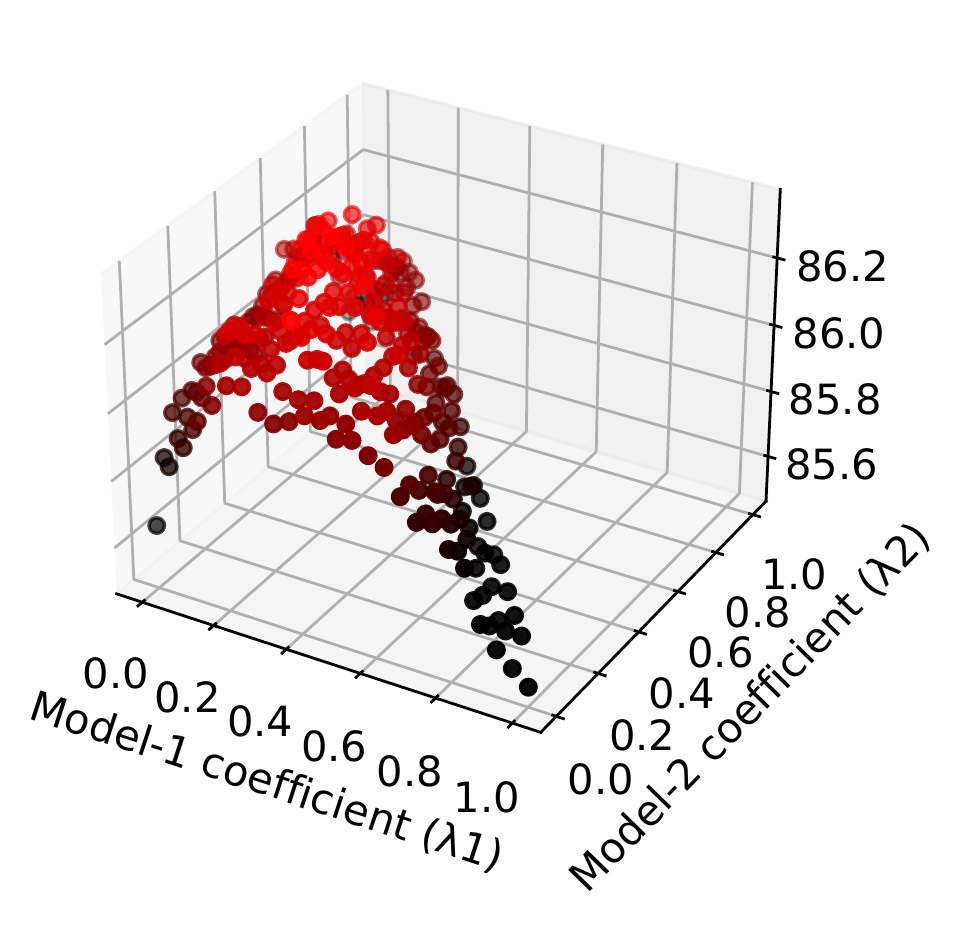}
        \caption{\textbf{Accuracy (\%) on interpolating the final converged models trained using DART (ours)} using WRN-28-10 model and CIFAR-100 dataset, by taking their convex combination. Maximum accuracy of $86.33$ is obtained on interpolating, using three experts with accuracies $85.65$, $85.75$ and $85.51$. For the best setting, $
        \lambda_1 = 0.17$ and $\lambda_2=0.46$.}
        \label{fig:interpolation_acc}
        \vspace{-0.2cm}
\end{figure}

While in the proposed method DART, we give equal weight to all $M$ branches, we note that fine-tuning the weights of individual models in a greedy manner \cite{wortsman2022model} can give a further boost in accuracy. As shown in Fig.\ref{fig:interpolation_acc}, the best accuracy obtained is $86.33\%$ at $\lambda_1 = 0.17$, $\lambda_2 = 0.46$, when compared to $86.24\%$ with  $\lambda_1 = \lambda_2 = 0.33$. These results are lower than those reported in Table-\ref{table:main_tab} of the main paper and Table-\ref{table:capacity} in the supplementary since the runs in  Fig.\ref{fig:interpolation_acc} do not use EMA, while our main method does.

\begin{figure*}
\centering
        \includegraphics[width=1\linewidth]{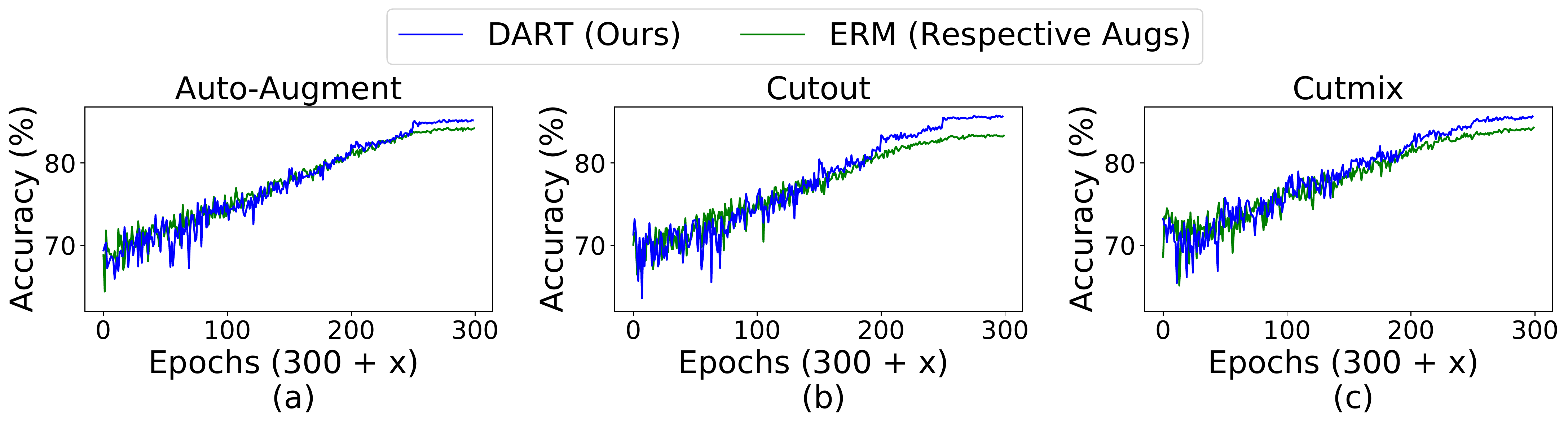}
        \caption{\textbf{Comparison of the test accuracy (\%) of ERM training using different augmentations with the respective augmentation expert of DART} on CIFAR-100, WRN-28-10. The analysis is done from 300 epochs onwards. Most of the gains of the proposed method occur at the end of training, when learning rate is low and the experts are present within a common basin.}
        \label{fig:gains_last}
\end{figure*}

\subsection{Training plots}

We show the training plots for In-domain generalization training of CIFAR-100 on WRN-28-10 in Fig.\ref{fig:gains_last}. We firstly note that not only does our method yield gains on the final interpolation step (as seen in Table-\ref{table:capacity} and Table-\ref{table:main_tab} of the main paper), but the step of intermediate interpolation ensures that the individual models are also better than the ERM baselines trained using the respective augmentations. Specifically, while the initial interpolations help in bringing the models closer to each other in the loss landscape, the later ones actually result in performance gains, since the low learning rate ensures that the flatter loss surface obtained using intermediate weight averaging is retained.

\vspace{-0.1cm}
\subsection{Integrating DART with SAM} Table-\ref{table:sam} shows that the proposed approach DART integrates effectively with SAM to obtain further performance gains. However, the gains are relatively lower on integrating with SAM ($\sim0.2\%$) when compared to the gains over Mixed ERM training ($\sim0.9\%$). We hypothesize that this is because SAM already encourages smoothness of loss surface, which is also achieved using DART. 

\input{tables/sam}

\vspace{-0.1cm}
\subsection{Evaluation across different model capacities}
\input{tables/capacity.tex}
We present results of DART on ResNet-18 and WideResNet-28-10 models in Table-\ref{table:capacity}. The gains obtained on WideResNet-28-10 are larger (0.2 and 0.89) when compared to ResNet-18 (0.06 and 0.64) demonstrating the scalability of our method.

\section{Details on Domain Generalization}

\subsection{Training Details}
\label{sec:train_details}
Since the domain shift across individual domains is larger in the Domain Generalization setting when compared to the In-Domain generalization setting, we found that training individual branches on a mix of all domains was better than training each branch on a single domain. Moreover, training on a mix of all domains also improves the individual branch accuracy, thereby boosting the accuracy of the final interpolated model. We train 4 branches (6 for DomainNet), where one branch is trained with an equal proportion of all domains, while the other three branches are allowed to be experts on individual domains by using a higher fraction (40\% for DomainNet and 50\% for other datasets) of the selected domain for the respective branch. 

In the Domain Generalization setting, the step of explicitly training on mixed augmentations / domains (L4-L5 in Algorithm-\ref{alg:WATIG} in the main paper) is replaced by the initialization of the model using ImageNet pretrained weights, which ensures that all models are in the loss basin. Moreover, this also helps in reducing the overall compute. 

\begin{figure}
\centering
        \includegraphics[width=1\linewidth]{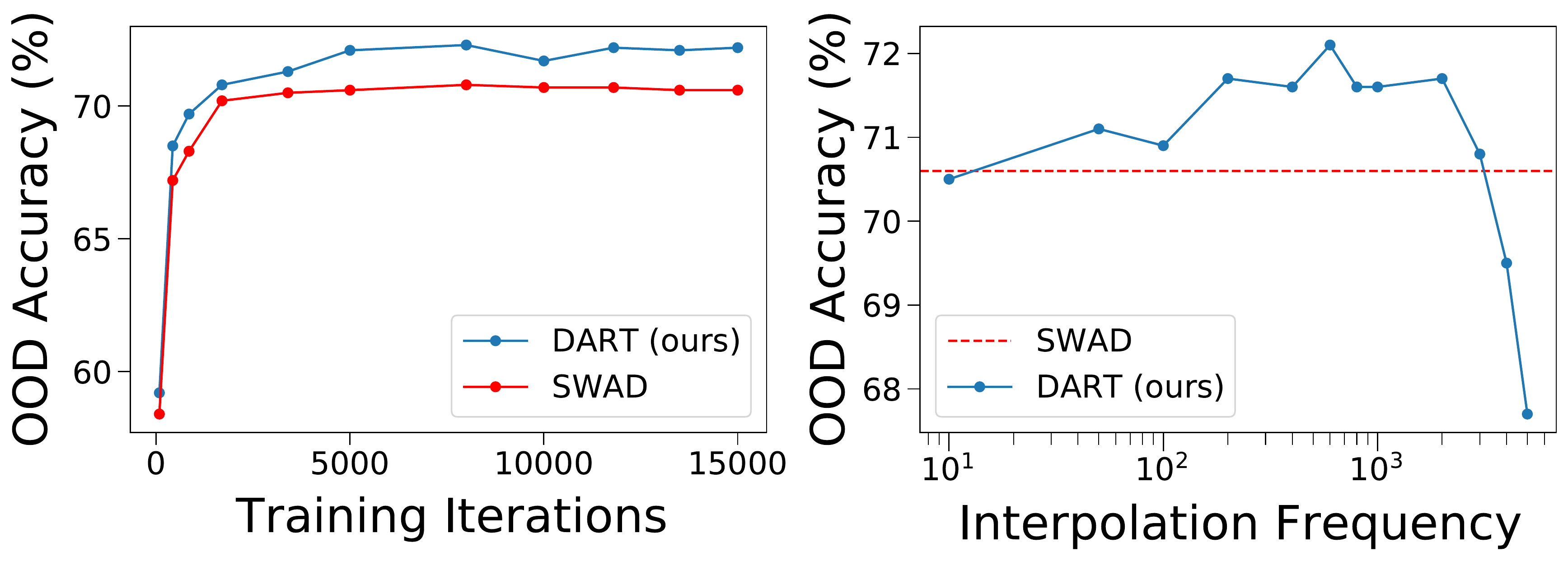}
         \vspace{-0.5cm}
         \caption{\textbf{Performance of DART across (a) varying training iterations and (b) varying interpolation frequency:} (a) compares the proposed method DART's performance with the SWAD baseline when trained for higher number of iterations. Interpolation frequency was maintained such that the number of interpolations remained same (8) in every case. (b) demonstrates the effect of intermediate interpolation frequency on DART. The training iterations were kept constant (5k).}
        \label{fig:dg_fig}
        \vspace{-0.2cm}
\end{figure}

For the results presented in the Tables \ref{tabular:avg} to \ref{tabular:domain} and Table-\ref{table:dg_cut} of the main paper, the training configuration (training iterations, interpolation frequency) was set to (15k, 1k) for DomainNet and (10k, 1k) for all other datasets, whereas for the results presented in Table-\ref{table:dg_2} in the main paper, the configuration was set to (5k, 600) for DANN \cite{ganin2016domain} and CDANN \cite{li2018domain}, and (8k, 1k) for the rest, primarily to reduce compute. The difference in adversarial training approaches (DANN and CDANN) was primarily because their training is not stable for longer training iterations. SWAD-specific hyperparameters were set as suggested by the authors \cite{cha2021swad} without additional tuning. While we compare with comparable compute for the In-Domain generalization setting (Table-\ref{table:capacity} and Table-\ref{table:main_tab} in the main paper), for the Domain Generalization setting we report the baselines from DomainBed \cite{gulrajani2020search} and the respective papers as is the common practice. Although the proposed approach uses higher compute than the baselines, we show in Fig.\ref{fig:dg_fig}(a) that even with higher compute, the baselines cannot achieve any better performance.

\subsection{Ablation experiments}

We present ablation experiments on the Office-Home dataset in Fig.\ref{fig:dg_fig}. Following this, we present average accuracy across all domain splits as is common practice in Domain Generalization \cite{gulrajani2020search}.

\textbf{Variation across training compute:} Fig.\ref{fig:dg_fig} (a) demonstrates that the performance of SWAD plateaus early compared to DART, when trained for a higher number of training iterations. We note that DART achieves a significant improvement in the final accuracy over the baseline. This indicates that although the proposed method requires higher compute, DART trades it off for improved performance.

\textbf{Variation in interpolation frequency:} Fig.\ref{fig:dg_fig} (b) describes the impact of varying the interpolation frequency in the proposed method DART. The number of training iterations is set to 5k for this experiment. We note that the accuracy is stable across a wide range of interpolation frequencies (x-axis is in log scale). This shows that the proposed method is not very sensitive to the frequency of interpolation, and does not require fine-tuning for every dataset. We therefore use the same interpolation frequency of 1k for all the datasets and training splits of DomainBed. We note that the proposed method performs better than baseline in all cases except when the frequency is kept too low ($\approx$10) or too high (close to total training iterations). The sharp deterioration in performance in the case of no intermediate interpolation (interpolation frequency = total training steps) illustrates the necessity of intermediate interpolation in the proposed method.

\begin{figure}
\centering
        \includegraphics[width=1\linewidth]{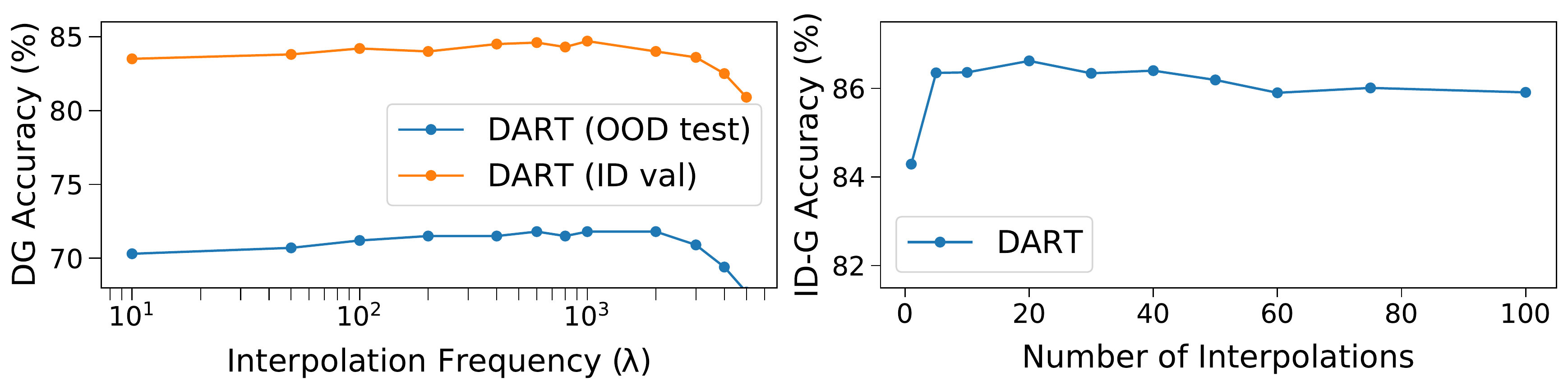}
         \caption{\textbf{Study on hyperparameter sensitivity for DART} (a) shows variation in the accuracy of DART for different interpolation frequencies on the OfficeHome dataset using ResNet-50 model with ImageNet initialization. A strong correlation between the in and out-of-domain accuracy of DART in the DG setting is observed. (b) shows the variation of In Domain Accuracy for the CIFAR-100 dataset and WRN-28-10 model vs. the number of interpolations done during the training. It is seen that the in-domain accuracy is stable across a wide range of interpolations. }
        \label{fig:hyper_fig}
\end{figure} 
\input{tables/DG_supple}

\end{document}

%% file: tables/motivation.tex
\begin{table}
\caption{\textbf{Motivation:} Performance (\%) on CIFAR100, ResNet-18 with ERM training for 200 epochs. Mixed-Training (MT) outperforms individual augmentations, and ensembles perform best.}
\vspace{-0.2cm}
\centering
\setlength\tabcolsep{3pt}
\resizebox{1.0\linewidth}{!}{
\label{table:motivation}
\begin{tabular}{lcccc}
\toprule
 & \multicolumn{4}{c}{Test Augmentation} \\
 \cmidrule{2-5}
Train Augmentation & No Aug. & Cutout & ~~~~Cutmix~~~~ & AutoAugment \\
\midrule
Pad+Crop+HFlip (PC) & 78.51 & 67.04 & 56.52 & 58.33 \\
Cutout (CO) & 77.99 & 74.58 & 56.12 & 58.47 \\
Cutmix (CM) & 80.54 & 74.05 & \textbf{77.35} & 61.23 \\
AutoAugment (AA) & 79.18 & 71.26 & 60.97 & 73.91 \\
Mixed-Training (MT) & 81.43 & 77.31 & 73.20 & \textbf{74.73} \\
\midrule
Ensemble (CM+CO+AA) & \textbf{83.61} & \textbf{79.19} & 73.19 & 73.90 \\
\bottomrule
\end{tabular}}
\vspace{-0.3cm}
\end{table}

%% file: algorithm/algo.tex
\begin{algorithm}[tb]
   \caption{Diversify-Aggregate-Repeat Training, DART}
   \label{alg:WATIG}
   
\begin{algorithmic}[1]
   \STATE {\bfseries Input:} $M$ networks $f_{\theta^{k}}$ where $0<k \leq M$, whose weights are aggregated every $\lambda$ epochs. Training Dataset for each network $f_{\theta^{k}}$ is represented by ${D^{k}} =\{(x^{k}_i,y^{k}_i) \}$. The union of all datasets is denoted as ${D^{*}}$. Number of training epochs E, Maximum Learning Rate $\mathrm{LR}_{max}$, Cross-entropy loss $\ell_{CE}$. Model is trained using ERM for $E^{'}$ epochs initially.

\FOR{$epoch=1$ {\bfseries to} $E$}
    \STATE $\mathrm{LR} = 0.5  \cdot \mathrm{LR}_{max} \cdot (1 + cos((epoch -1)/E\cdot \pi))$
    \IF{ $epoch<E'$}
    \STATE $\theta = min_{\theta}\frac{1}{n}\sum\limits_{i=1}^n\ell_{CE}(\theta,{D^{*}})$ 
        \ELSE
        \IF{$epoch=E'$}
   \STATE $\theta^{k}\leftarrow \theta ~~ \forall k\in [1,M]$ 
        \ENDIF
      \STATE  $\theta^{k} = min_{\theta^{k}}\frac{1}{n}\sum\limits_{i=1}^n\ell_{CE}(\theta,{D^{k}})$ $\forall k \in [1,M]$ 
    
\IF{ $epoch ~ \%~\lambda=0$}
    \STATE $\theta = \frac{1}{M}\sum\limits_{k=1}^M\theta^{k}$ 
   \STATE $\theta^{k}\leftarrow \theta ~~ \forall k\in [1,M]$ 
\ENDIF
\ENDIF
\ENDFOR
\end{algorithmic}
\end{algorithm}

%% file: tables/ID.tex
\begin{table}
\caption{\textbf{In-Domain Generalization:} Performance (\%) of DART when compared to baselines on WideResNet-28-10 model. Standard deviation for DART and MT is reported across 5 reruns.}
\setlength\tabcolsep{2pt}
\resizebox{1.0\linewidth}{!}{
\label{table:main_tab}
\begin{tabular}{lll}
\toprule
Method & CIFAR-10 ~~~~~~~& CIFAR-100~~~~~~~
     \\
\midrule
ERM+EMA (Pad+Crop+HFlip) & 96.41  & 81.67  \\
ERM+EMA (AutoAugment) & 97.50 & 84.20   \\
ERM+EMA (Cutout) & 97.43 & 82.33  
 \\
ERM+EMA (Cutmix)  & 97.11 & 84.05  
 \\
Learning Subspaces \cite{wortsman21alearningsubspace} 
& 97.46  & 83.91   \\
\midrule
ERM+EMA (Mixed Training-MT)~~~~~~~~~ & 97.69 \std{0.19} & 85.57 \std{0.13}           \\
DART (Ours) & \textbf{97.96} \std{0.06} & \textbf{86.46} \std{0.12}  \\
\bottomrule
\end{tabular}}
\vspace{-0.2cm}
\end{table}

%% file: tables/Imagenet.tex
\begin{table}
\caption{\textbf{DART on ImageNet-1K and finegrained datasets:} Performance ($\%$) of DART when compared to ERM+EMA Mixed Training baseline on ResNet-50. In the first row, a \textit{Single Augmentation (SA)} is used in all branches (RandAugment \cite{cubuk2020randaugment} for ImageNet-1K, and Pad-Crop for finegrained datasets). In the second row, \textit{Mixed Augmentations (MA)} - Pad-Crop, RandAugment \cite{cubuk2020randaugment} and Cutout \cite{devries2017improved} are used in different branches. AutoAugment \cite{cubuk2018autoaugment} is used instead of RandAugment for finegrained datasets in the latter case of Mixed Augmentations (MA).}
\vspace{-0.1cm}
\centering
\resizebox{1.0\linewidth}{!}{
\label{table:imagenet}
\begin{tabular}{lcccccc}
\toprule

             & \multicolumn{2}{c}{\textbf{\small{Stanford-CARS}}}                                         & \multicolumn{2}{c}{\small{\textbf{CUB-200}}}                                                         & \multicolumn{2}{c}{\textbf{\small{Imagenet-1K}}} \\
 & \multicolumn{1}{c}{\textbf{ERM + EMA}} & \multicolumn{1}{c}{\textbf{DART}} & \multicolumn{1}{c}{\textbf{ERM + EMA}} & \multicolumn{1}{c}{\textbf{DART}} &\textbf{ERM + EMA}       & \textbf{DART}               \\
\midrule
SA& 88.11                               & \textbf{90.42}                             & 78.55                       & \textbf{79.75}                     & 78.55      & \textbf{78.96}     \\
MA & 90.88                               & \textbf{91.95}                             & 81.72                       & \textbf{82.83}                     & 79.06      & \textbf{79.20}  \\

\bottomrule
\end{tabular}
}
\vspace{-0.3cm}
\end{table}

%% file: tables/DG_cut.tex
\begin{table}
\caption{\textbf{Domain Generalization:} OOD accuracy(\%) of DART when compared to the respective baselines on DomainBed datasets with ResNet-50 model. Standard dev. across 3 reruns is reported.}
\vspace{-0.5cm}
\label{table:dg_cut}
\begin{center}
\resizebox{1.0\linewidth}{!}{%
\begin{tabular}{lcccccc}
\toprule
\textbf{Algorithm}& \textbf{VLCS}& \textbf{PACS}& \textbf{OfficeHome}& \textbf{TerraInc}   & \textbf{DomainNet}& \textbf{Avg}\\
\midrule
ERM \cite{vapnik1998statistical}
& 77.5 $\pm$ 0.4    & 85.5 $\pm$ 0.2    & 66.5 $\pm$ 0.3
& 46.1 $\pm$ 1.8    & 40.9 $\pm$ 0.1    & 63.3\\
+ DART \small (Ours) 
& 78.5 $\pm$ 0.7    & 87.3 $\pm$ 0.5    & 70.1 $\pm$ 0.2 
& 48.7 $\pm$ 0.8    & 45.8 $\pm$ 0.0    & 66.1       \\
\midrule
SWAD     \cite{cha2021swad}
& 79.1 $\pm$ 0.1    & 88.1 $\pm$ 0.1    & 70.6 $\pm$ 0.2 
& 50.0 $\pm$ 0.3    & 46.5 $\pm$ 0.1    & 66.9       \\
+ DART \small (Ours)
& \textbf{80.3} $\pm$ 0.2    & \textbf{88.9} $\pm$ 0.1    & \textbf{71.9} $\pm$ 0.1 
& \textbf{51.3} $\pm$ 0.2    & \textbf{47.1} $\pm$ 0.0    & \textbf{67.9}     \\
\bottomrule

\end{tabular}}
\vspace{-0.4cm}
\end{center}
\end{table}

%% file: tables/DG_algos.tex
\begin{table}
\caption{\textbf{Combining DART with other DG methods (OfficeHome):} OOD performance (\%) of the proposed method DART coupled with different algorithms against their vanilla and SWAD counterparts. Numbers represented with $^\dagger$ were reproduced while others are from Domainbed \cite{gulrajani2020search}. All models except the last row are trained on a ResNet-50 Imagenet pretrained model. The last row shows results on a CLIP initialized ViT-B/16 model.}
\vspace{-0.5cm}
\label{table:dg_2}
\begin{center}
\setlength\tabcolsep{3pt}
\resizebox{1.0\linewidth}{!}{%
\begin{tabular}{lcccc}
\toprule
\textbf{Algorithm}& \textbf{Vanilla}& \textbf{DART} (\small w/o SWAD) & \textbf{SWAD}& \textbf{DART} (\small + SWAD) \\
\midrule
ERM \cite{vapnik1998statistical}
& 66.5    & 70.31     & 70.60
& \textbf{72.28}    \\
ARM      \cite{zhang2021adaptive}         
& 64.8    & 69.24    & 69.75
& \textbf{71.31}    \\
SAM$^{\dagger}$ \cite{foret2020sharpness}
& 67.4    & 70.39    & 70.26 
& \textbf{71.55}    \\
Cutmix$^{\dagger}$ \cite{yun2019cutmix}         
& 67.3    & 70.07    & 71.08
& \textbf{71.49}   \\
Mixup      \cite{wang2020heterogeneous}                    
& 68.1    & 71.14    & 71.15   
& \textbf{72.38}    \\
DANN     \cite{ganin2016domain}                   
& 65.9    & 70.32    & 69.46
& \textbf{70.85}   \\
CDANN    \cite{li2018domain}                      
& 65.8    & 70.75    & 69.70
& \textbf{71.69}    \\
SagNet   \cite{nam2021reducing}                
& 68.1    & 70.19    & 70.84 
& \textbf{71.96}   \\
MIRO \cite{cha2022miro}                
& 70.5    & 72.54    & 72.40 
& \textbf{72.71}   \\
MIRO (CLIP)$^\dagger$
& 83.3 & 86.14 & 84.80 & \textbf{87.37} \\
\bottomrule
\end{tabular}}
\end{center}
\vspace{-0.7cm}
\end{table}

%% file: tables/augs_impact.tex
\begin{table}
\caption{\textbf{DART using same augmentation across all branches:} Performance (\%) of DART when compared to baselines across different augmentations on CIFAR-100 using WideResNet-28-10 architecture. DART is better than baselines in all cases.}
\setlength\tabcolsep{3pt}
\resizebox{1.0\linewidth}{!}{
\label{table:augs}
\begin{tabular}{lccccc}
\toprule

Method & Pad+Crop+HFlip & AutoAug. & Cutout & Cutmix & Mixed-Train. \\ 
\midrule
ERM & 81.48  & 83.93 & 82.01 & 83.02 & 85.54  \\
ERM + EMA & 81.67  & 84.20           & 82.33            & 84.05            & 85.57   \\
DART (Ours)                                                             & \textbf{82.31}                                                        & \textbf{85.02}   & \textbf{84.15}     & \textbf{84.72}    & \textbf{86.13}   \\

\bottomrule
\end{tabular}
}
\vspace{-0.5cm}
\end{table}

%% file: tables/loss_landscape.tex
\begin{table}
\caption{\textbf{Loss Landscape Sharpness Analysis:} Comparison of the proposed method DART (Pad-Crop) and ERM (Pad-Crop) trained using WRN-28-10 on CIFAR-100. The metrics presented here have been adapted from Stutz \etal \cite{stutz2021relating}. For all metrics, a lower value corresponds to a flatter loss landscape.}
\setlength\tabcolsep{2pt}
\resizebox{1.0\linewidth}{!}{
\label{table:loss_landscape}
\begin{tabular}{lccc}
\toprule        
\textbf{Method} &\textbf{Worst  Case}            &\textbf{Average} & \textbf{Average} \\
  & \textbf{Flatness $\downarrow$} & \textbf{Flatness $\downarrow$} & \textbf{Train Loss $\downarrow$} \\
\midrule
ERM (Pad+Crop) & 4.173 & 1.090                     & 0.0028                      \\ 
DART (Pad+Crop) (Ours)                                                & \textbf{2.037}                                                       & \textbf{0.294}                     & \textbf{0.0022}   \\
\bottomrule
\end{tabular}}
\vspace{-0.4cm}
\end{table}

%% file: tables/sam.tex
\begin{table}
\caption{\textbf{Integrating DART with SAM} gives around 0.2\% improvement in performance (\%) when compared to SAM with mixed augmentations. The results are shown on CIFAR-100 dataset using WRN-28-10 model.}
\setlength\tabcolsep{3pt}
\resizebox{1.0\linewidth}{!}{
\label{table:sam}

\begin{tabular}{ccccc}
\toprule

                                    ERM+EMA & ERM+SWA  & DART & SAM+EMA & DART+SAM+EMA   \\
                                       \midrule
 85.57 \std{0.13}       & 85.44 \std{0.09}       & 86.46 \std{0.12}  & 87.05 \std{0.15}      & \textbf{87.26} \std{0.02} \\

\bottomrule
\end{tabular}
}
\end{table}

%% file: tables/capacity.tex
\begin{table}
\caption{\textbf{Different model architectures:} Performance (\%) of DART when compared to Mixed-Training (MT) across different architectures. Standard deviation is reported across 5 reruns.}
\vspace{-0.15cm}
\centering
\setlength\tabcolsep{3pt}
\resizebox{1.0\linewidth}{!}{
\label{table:capacity}
\begin{tabular}{llcc}
\toprule
Model & Method & ~~CIFAR-10~~ & CIFAR-100    
     \\
\midrule
ResNet18 & ERM+EMA (Mixed - MT) & 97.08 \std{0.05} & 82.25 \std{0.29} \\
 & DART (Ours) & \textbf{97.14} \std{0.08} & \textbf{82.89} \std{0.07} \\ 
 \midrule
 WRN-28-10 & ERM+EMA (Mixed - MT)~~~~ & 97.76 \std{0.17}  & 85.57 \std{0.13} \\
 & DART (Ours) & \textbf{97.96} \std{0.06} & \textbf{86.46} \std{0.12} \\ 
\bottomrule
\end{tabular}}
\end{table}

%% file: tables/DG_supple.tex
\subsection{Detailed Results}
\vspace{-0.1cm}
In this section, we present complete Domain Generalization results (Out-of-domain accuracies in \%) on VLCS (Table-\ref{tabular:vlcs}), PACS (Table-\ref{tabular:pacs}), OfficeHome (Table-\ref{tabular:office}), TerraIncognita (Table-\ref{tabular:terra}) and DomainNet (Table-\ref{tabular:domain}) benchmarks. We also present the average accuracy across all domain splits and datasets in Table-\ref{tabular:avg}. We note that the proposed method DART when combined with SWAD \cite{cha2021swad} outperforms all existing methods across all datasets.  

\subsubsection{Averages}
\vspace{-0.1cm}
\begin{center}
\resizebox{1.0\linewidth}{!}{%
\begin{tabular}{lcccccc}
\toprule
\textbf{Algorithm}& \textbf{VLCS}& \textbf{PACS}& \textbf{OfficeHome}& \textbf{TerraIncognita}   & \textbf{DomainNet}& \textbf{Avg}\\
\midrule
ERM \cite{vapnik1998statistical}
& 77.5 $\pm$ 0.4    & 85.5 $\pm$ 0.2    & 66.5 $\pm$ 0.3
& 46.1 $\pm$ 1.8    & 40.9 $\pm$ 0.1    & 63.3\\
IRM  \cite{arjovsky2019invariant}         
& 78.5 $\pm$ 0.5    & 83.5 $\pm$ 0.8    & 64.3 $\pm$ 2.2
& 47.6 $\pm$ 0.8    & 33.9 $\pm$ 2.8    & 61.6\\
GroupDRO   \cite{sagawa2019distributionally}
& 76.7 $\pm$ 0.6    & 84.4 $\pm$ 0.8    & 66.0 $\pm$ 0.7 
& 43.2 $\pm$ 1.1    & 33.3 $\pm$ 0.2    & 60.7  \\
Mixup      \cite{wang2020heterogeneous}
& 77.4 $\pm$ 0.6    & 84.6 $\pm$ 0.6    & 68.1 $\pm$ 0.3
& 47.9 $\pm$ 0.8    & 39.2 $\pm$ 0.1    & 63.4    \\
MLDG    \cite{li2018learning}         
& 77.2 $\pm$ 0.4    & 84.9 $\pm$ 1.0    & 66.8 $\pm$ 0.6
& 47.7 $\pm$ 0.9    & 41.2 $\pm$ 0.1    & 63.6     \\
CORAL    \cite{sun2016deep}                      
& 78.8 $\pm$ 0.6    & 86.2 $\pm$ 0.3    & 68.7 $\pm$ 0.3
& 47.6 $\pm$ 1.0    & 41.5 $\pm$ 0.1    & 64.5      \\
MMD     \cite{8578664}                          
& 77.5 $\pm$ 0.9    & 84.6 $\pm$ 0.5    & 66.3 $\pm$ 0.1   
& 42.2 $\pm$ 1.6    & 23.4 $\pm$ 9.5    & 58.8       \\
DANN     \cite{ganin2016domain}                   
& 78.6 $\pm$ 0.4    & 83.6 $\pm$ 0.4    & 65.9 $\pm$ 0.6 
& 46.7 $\pm$ 0.5    & 38.3 $\pm$ 0.1    & 62.6       \\
CDANN    \cite{li2018domain}                      
& 77.5 $\pm$ 0.1    & 82.6 $\pm$ 0.9    & 65.8 $\pm$ 1.3
& 45.8 $\pm$ 1.6    & 38.3 $\pm$ 0.3    & 62.0       \\
MTL     \cite{blanchard2021domain}                
& 77.2 $\pm$ 0.4    & 84.6 $\pm$ 0.5    & 66.4 $\pm$ 0.5 
& 45.6 $\pm$ 1.2    & 40.6 $\pm$ 0.1    & 62.9       \\
SagNet   \cite{nam2021reducing}                     
& 77.8 $\pm$ 0.5    & 86.3 $\pm$ 0.2    & 68.1 $\pm$ 0.1
& 48.6 $\pm$ 1.0    & 40.3 $\pm$ 0.1    & 64.2       \\
ARM      \cite{zhang2021adaptive}        
& 77.6 $\pm$ 0.3    & 85.1 $\pm$ 0.4    & 64.8 $\pm$ 0.3   
& 45.5 $\pm$ 0.3    & 35.5 $\pm$ 0.2    & 61.7       \\
VREx     \cite{krueger2021out}       
& 78.3 $\pm$ 0.2    & 84.9 $\pm$ 0.6    & 66.4 $\pm$ 0.6
& 46.4 $\pm$ 0.6    & 33.6 $\pm$ 2.9    & 61.9       \\
RSC      \cite{huang2020self}                     
& 77.1 $\pm$ 0.5    & 85.2 $\pm$ 0.9    & 65.5 $\pm$ 0.9 
& 46.6 $\pm$ 1.0    & 38.9 $\pm$ 0.5    & 62.7       \\
SWAD     \cite{cha2021swad}
& 79.1 $\pm$ 0.1    & 88.1 $\pm$ 0.1    & 70.6 $\pm$ 0.2 
& 50.0 $\pm$ 0.3    & 46.5 $\pm$ 0.1    & 66.9       \\
\midrule
DART \small w/o SWAD 
& 78.5 $\pm$ 0.7    & 87.3 $\pm$ 0.5    & 70.1 $\pm$ 0.2 
& 48.7 $\pm$ 0.8    & 45.8              & 66.1       \\
DART \small w/ SWAD
& \textbf{80.3} $\pm$ 0.2    & \textbf{88.9} $\pm$ 0.1    & \textbf{71.9} $\pm$ 0.1 
& \textbf{51.3} $\pm$ 0.2    & \textbf{47.2}              & \textbf{67.9}       \\
\bottomrule
\label{tabular:avg}
\end{tabular}}
\end{center}
\subsubsection{VLCS}
\vspace{-0.3cm}
\begin{center}
\resizebox{1.0\linewidth}{!}{%
\begin{tabular}{lccccc}
\toprule
\textbf{Algorithm}   & \textbf{C}           & \textbf{L}           & \textbf{S}           & \textbf{V}           & \textbf{Avg}         \\
\midrule
ERM                  & 97.7 $\pm$ 0.4       & 64.3 $\pm$ 0.9       & 73.4 $\pm$ 0.5       & 74.6 $\pm$ 1.3       & 77.5                 \\
IRM                  & 98.6 $\pm$ 0.1       & 64.9 $\pm$ 0.9       & 73.4 $\pm$ 0.6       & 77.3 $\pm$ 0.9       & 78.5                 \\
GroupDRO             & 97.3 $\pm$ 0.3       & 63.4 $\pm$ 0.9       & 69.5 $\pm$ 0.8       & 76.7 $\pm$ 0.7       & 76.7                 \\
Mixup                & 98.3 $\pm$ 0.6       & 64.8 $\pm$ 1.0       & 72.1 $\pm$ 0.5       & 74.3 $\pm$ 0.8       & 77.4                 \\
MLDG                 & 97.4 $\pm$ 0.2       & 65.2 $\pm$ 0.7       & 71.0 $\pm$ 1.4       & 75.3 $\pm$ 1.0       & 77.2                 \\
CORAL                & 98.3 $\pm$ 0.1       & 66.1 $\pm$ 1.2       & 73.4 $\pm$ 0.3       & 77.5 $\pm$ 1.2       & 78.8                 \\
MMD                  & 97.7 $\pm$ 0.1       & 64.0 $\pm$ 1.1       & 72.8 $\pm$ 0.2       & 75.3 $\pm$ 3.3       & 77.5                 \\
DANN                 & \textbf{99.0} $\pm$ 0.3       & 65.1 $\pm$ 1.4       & 73.1 $\pm$ 0.3       & 77.2 $\pm$ 0.6       & 78.6                 \\
CDANN                & 97.1 $\pm$ 0.3       & 65.1 $\pm$ 1.2       & 70.7 $\pm$ 0.8       & 77.1 $\pm$ 1.5       & 77.5                 \\
MTL                  & 97.8 $\pm$ 0.4       & 64.3 $\pm$ 0.3       & 71.5 $\pm$ 0.7       & 75.3 $\pm$ 1.7       & 77.2                 \\
SagNet               & 97.9 $\pm$ 0.4       & 64.5 $\pm$ 0.5       & 71.4 $\pm$ 1.3       & 77.5 $\pm$ 0.5       & 77.8                 \\
ARM                  & 98.7 $\pm$ 0.2       & 63.6 $\pm$ 0.7       & 71.3 $\pm$ 1.2       & 76.7 $\pm$ 0.6       & 77.6                 \\
VREx                 & 98.4 $\pm$ 0.3       & 64.4 $\pm$ 1.4       & 74.1 $\pm$ 0.4       & 76.2 $\pm$ 1.3       & 78.3                 \\
RSC                  & 97.9 $\pm$ 0.1       & 62.5 $\pm$ 0.7       & 72.3 $\pm$ 1.2       & 75.6 $\pm$ 0.8       & 77.1                 \\
SWAD                 & 98.8 $\pm$ 0.1       & 63.3 $\pm$ 0.3       & 75.3 $\pm$ 0.5       & 79.2 $\pm$ 0.6       & 79.1                 \\
\midrule
DART \small w/o SWAD & 97.9 $\pm$ 1.0       & 64.2 $\pm$ 0.7       & 73.9 $\pm$ 1.1       & 78.1 $\pm$ 1.6       & 78.5                 \\
DART \small w/ SWAD  & 98.7 $\pm$ 0.0       & \textbf{66.4} $\pm$ 0.3       & \textbf{75.8} $\pm$ 0.6       & \textbf{80.4} $\pm$ 0.3       & \textbf{80.3}                 \\
\bottomrule
\label{tabular:vlcs}
\end{tabular}}
\end{center}

\subsubsection{PACS}
\vspace{-0.3cm}
\begin{center}
\resizebox{1.0\linewidth}{!}{%
\begin{tabular}{lccccc}
\toprule
\textbf{Algorithm}   & \textbf{A}           & \textbf{C}           & \textbf{P}           & \textbf{S}           & \textbf{Avg}         \\
\midrule
ERM                  & 84.7 $\pm$ 0.4       & 80.8 $\pm$ 0.6       & 97.2 $\pm$ 0.3       & 79.3 $\pm$ 1.0       & 85.5                 \\
IRM                  & 84.8 $\pm$ 1.3       & 76.4 $\pm$ 1.1       & 96.7 $\pm$ 0.6       & 76.1 $\pm$ 1.0       & 83.5                 \\
GroupDRO             & 83.5 $\pm$ 0.9       & 79.1 $\pm$ 0.6       & 96.7 $\pm$ 0.3       & 78.3 $\pm$ 2.0       & 84.4                 \\
Mixup                & 86.1 $\pm$ 0.5       & 78.9 $\pm$ 0.8       & 97.6 $\pm$ 0.1       & 75.8 $\pm$ 1.8       & 84.6                 \\
MLDG                 & 85.5 $\pm$ 1.4       & 80.1 $\pm$ 1.7       & 97.4 $\pm$ 0.3       & 76.6 $\pm$ 1.1       & 84.9                 \\
CORAL                & 88.3 $\pm$ 0.2       & 80.0 $\pm$ 0.5       & 97.5 $\pm$ 0.3       & 78.8 $\pm$ 1.3       & 86.2                 \\
MMD                  & 86.1 $\pm$ 1.4       & 79.4 $\pm$ 0.9       & 96.6 $\pm$ 0.2       & 76.5 $\pm$ 0.5       & 84.6                 \\
DANN                 & 86.4 $\pm$ 0.8       & 77.4 $\pm$ 0.8       & 97.3 $\pm$ 0.4       & 73.5 $\pm$ 2.3       & 83.6                 \\
CDANN                & 84.6 $\pm$ 1.8       & 75.5 $\pm$ 0.9       & 96.8 $\pm$ 0.3       & 73.5 $\pm$ 0.6       & 82.6                 \\
MTL                  & 87.5 $\pm$ 0.8       & 77.1 $\pm$ 0.5       & 96.4 $\pm$ 0.8       & 77.3 $\pm$ 1.8       & 84.6                 \\
SagNet               & 87.4 $\pm$ 1.0       & 80.7 $\pm$ 0.6       & 97.1 $\pm$ 0.1       & 80.0 $\pm$ 0.4       & 86.3                 \\
ARM                  & 86.8 $\pm$ 0.6       & 76.8 $\pm$ 0.5       & 97.4 $\pm$ 0.3       & 79.3 $\pm$ 1.2       & 85.1                 \\
VREx                 & 86.0 $\pm$ 1.6       & 79.1 $\pm$ 0.6       & 96.9 $\pm$ 0.5       & 77.7 $\pm$ 1.7       & 84.9                 \\
RSC                  & 85.4 $\pm$ 0.8       & 79.7 $\pm$ 1.8       & 97.6 $\pm$ 0.3       & 78.2 $\pm$ 1.2       & 85.2                 \\
DMG \cite{chattopadhyay2020learning}                 & 82.6                 & 78.1                 & 94.3                 & 78.3                 & 83.4                 \\
MetaReg \cite{NEURIPS2018_647bba34}             & 87.2                 & 79.2                 & 97.6                 & 70.3                 & 83.6                 \\
DSON  \cite{dson}               & 87.0                 & 80.6                 & 96.0                 & 82.9                 & 86.6                 \\
SWAD                 & 89.3 $\pm$ 0.2       & 83.4 $\pm$ 0.6       & 97.3 $\pm$ 0.3       & 78.2 $\pm$ 0.5       & 88.1                 \\
\midrule
DART \small w/o SWAD & 87.1 $\pm$ 1.5       & 83.5 $\pm$ 0.9       & 96.9 $\pm$ 0.3       & 81.8 $\pm$ 0.9       & 87.3                 \\
DART \small w/ SWAD  & \textbf{90.1} $\pm$ 0.1       & \textbf{84.5} $\pm$ 0.2       & \textbf{97.7} $\pm$ 0.2       & \textbf{83.4} $\pm$ 0.1       & \textbf{88.9}                 \\
\bottomrule
\label{tabular:pacs}
\end{tabular}}
\end{center}

\subsubsection{OfficeHome}
\vspace{-0.3cm}
\begin{center}
\resizebox{1.0\linewidth}{!}{%
\begin{tabular}{lccccc}
\toprule
\textbf{Algorithm}   & \textbf{A}           & \textbf{C}           & \textbf{P}           & \textbf{R}           & \textbf{Avg}         \\
\midrule
ERM                  & 61.3 $\pm$ 0.7       & 52.4 $\pm$ 0.3       & 75.8 $\pm$ 0.1       & 76.6 $\pm$ 0.3       & 66.5                 \\
IRM                  & 58.9 $\pm$ 2.3       & 52.2 $\pm$ 1.6       & 72.1 $\pm$ 2.9       & 74.0 $\pm$ 2.5       & 64.3                 \\
GroupDRO             & 60.4 $\pm$ 0.7       & 52.7 $\pm$ 1.0       & 75.0 $\pm$ 0.7       & 76.0 $\pm$ 0.7       & 66.0                 \\
Mixup                & 62.4 $\pm$ 0.8       & 54.8 $\pm$ 0.6       & 76.9 $\pm$ 0.3       & 78.3 $\pm$ 0.2       & 68.1                 \\
MLDG                 & 61.5 $\pm$ 0.9       & 53.2 $\pm$ 0.6       & 75.0 $\pm$ 1.2       & 77.5 $\pm$ 0.4       & 66.8                 \\
CORAL                & 65.3 $\pm$ 0.4       & 54.4 $\pm$ 0.5       & 76.5 $\pm$ 0.1       & 78.4 $\pm$ 0.5       & 68.7                 \\
MMD                  & 60.4 $\pm$ 0.2       & 53.3 $\pm$ 0.3       & 74.3 $\pm$ 0.1       & 77.4 $\pm$ 0.6       & 66.3                 \\
DANN                 & 59.9 $\pm$ 1.3       & 53.0 $\pm$ 0.3       & 73.6 $\pm$ 0.7       & 76.9 $\pm$ 0.5       & 65.9                 \\
CDANN                & 61.5 $\pm$ 1.4       & 50.4 $\pm$ 2.4       & 74.4 $\pm$ 0.9       & 76.6 $\pm$ 0.8       & 65.8                 \\
MTL                  & 61.5 $\pm$ 0.7       & 52.4 $\pm$ 0.6       & 74.9 $\pm$ 0.4       & 76.8 $\pm$ 0.4       & 66.4                 \\
SagNet               & 63.4 $\pm$ 0.2       & 54.8 $\pm$ 0.4       & 75.8 $\pm$ 0.4       & 78.3 $\pm$ 0.3       & 68.1                 \\
ARM                  & 58.9 $\pm$ 0.8       & 51.0 $\pm$ 0.5       & 74.1 $\pm$ 0.1       & 75.2 $\pm$ 0.3       & 64.8                 \\
VREx                 & 60.7 $\pm$ 0.9       & 53.0 $\pm$ 0.9       & 75.3 $\pm$ 0.1       & 76.6 $\pm$ 0.5       & 66.4                 \\
RSC                  & 60.7 $\pm$ 1.4       & 51.4 $\pm$ 0.3       & 74.8 $\pm$ 1.1       & 75.1 $\pm$ 1.3       & 65.5                 \\
SWAD                 & 66.1 $\pm$ 0.4       & 57.7 $\pm$ 0.4       & 78.4 $\pm$ 0.1       & 80.2 $\pm$ 0.2       & 70.6                 \\
\midrule
DART \small w/o SWAD & 64.3 $\pm$ 0.2       & 57.9 $\pm$ 0.9       & 78.3 $\pm$ 0.6       & 79.9 $\pm$ 0.1       & 70.1                 \\
DART \small w/ SWAD  & \textbf{67.1} $\pm$ 0.2       & \textbf{59.2} $\pm$ 0.1       & \textbf{79.7} $\pm$ 0.1       & \textbf{81.5} $\pm$ 0.1       & \textbf{71.9}                 \\
\bottomrule
\label{tabular:office}
\end{tabular}}
\end{center}

\subsubsection{TerraIncognita}
\vspace{-0.3cm}
\begin{center}
\resizebox{1.0\linewidth}{!}{%
\begin{tabular}{lccccc}
\toprule
\textbf{Algorithm}   & \textbf{L100}        & \textbf{L38}         & \textbf{L43}         & \textbf{L46}         & \textbf{Avg}         \\
\midrule
ERM                  & 49.8 $\pm$ 4.4       & 42.1 $\pm$ 1.4       & 56.9 $\pm$ 1.8       & 35.7 $\pm$ 3.9       & 46.1                 \\
IRM                  & 54.6 $\pm$ 1.3       & 39.8 $\pm$ 1.9       & 56.2 $\pm$ 1.8       & 39.6 $\pm$ 0.8       & 47.6                 \\
GroupDRO             & 41.2 $\pm$ 0.7       & 38.6 $\pm$ 2.1       & 56.7 $\pm$ 0.9       & 36.4 $\pm$ 2.1       & 43.2                 \\
Mixup                & 59.6 $\pm$ 2.0       & 42.2 $\pm$ 1.4       & 55.9 $\pm$ 0.8       & 33.9 $\pm$ 1.4       & 47.9                 \\
MLDG                 & 54.2 $\pm$ 3.0       & 44.3 $\pm$ 1.1       & 55.6 $\pm$ 0.3       & 36.9 $\pm$ 2.2       & 47.7                 \\
CORAL                & 51.6 $\pm$ 2.4       & 42.2 $\pm$ 1.0       & 57.0 $\pm$ 1.0       & 39.8 $\pm$ 2.9       & 47.6                 \\
MMD                  & 41.9 $\pm$ 3.0       & 34.8 $\pm$ 1.0       & 57.0 $\pm$ 1.9       & 35.2 $\pm$ 1.8       & 42.2                 \\
DANN                 & 51.1 $\pm$ 3.5       & 40.6 $\pm$ 0.6       & 57.4 $\pm$ 0.5       & 37.7 $\pm$ 1.8       & 46.7                 \\
CDANN                & 47.0 $\pm$ 1.9       & 41.3 $\pm$ 4.8       & 54.9 $\pm$ 1.7       & 39.8 $\pm$ 2.3       & 45.8                 \\
MTL                  & 49.3 $\pm$ 1.2       & 39.6 $\pm$ 6.3       & 55.6 $\pm$ 1.1       & 37.8 $\pm$ 0.8       & 45.6                 \\
SagNet               & 53.0 $\pm$ 2.9       & 43.0 $\pm$ 2.5       & 57.9 $\pm$ 0.6       & 40.4 $\pm$ 1.3       & 48.6                 \\
ARM                  & 49.3 $\pm$ 0.7       & 38.3 $\pm$ 2.4       & 55.8 $\pm$ 0.8       & 38.7 $\pm$ 1.3       & 45.5                 \\
VREx                 & 48.2 $\pm$ 4.3       & 41.7 $\pm$ 1.3       & 56.8 $\pm$ 0.8       & 38.7 $\pm$ 3.1       & 46.4                 \\
RSC                  & 50.2 $\pm$ 2.2       & 39.2 $\pm$ 1.4       & 56.3 $\pm$ 1.4       & \textbf{40.8} $\pm$ 0.6       & 46.6                 \\
SWAD                 & 55.4 $\pm$ 0.0       & 44.9 $\pm$ 1.1       & 59.7 $\pm$ 0.4       & 39.9 $\pm$ 0.2       & 50.0                 \\
\midrule
DART \small w/o SWAD & 54.6 $\pm$ 1.1       & 44.9 $\pm$ 1.6       & 58.7 $\pm$ 0.5       & 36.6 $\pm$ 1.9       & 48.7                 \\
DART \small w/ SWAD  & \textbf{56.3} $\pm$ 0.4       & \textbf{47.1} $\pm$ 0.3       & \textbf{61.2} $\pm$ 0.3       & 40.5 $\pm$ 0.1       & \textbf{51.3}                 \\
\bottomrule
\label{tabular:terra}
\end{tabular}}
\end{center}

\subsubsection{DomainNet}
\vspace{-0.3cm}
\begin{center}
\resizebox{1.0\linewidth}{!}{%
\begin{tabular}{lccccccc}
\toprule
\textbf{Algorithm}   & \textbf{clip}        & \textbf{info}        & \textbf{paint}       & \textbf{quick}       & \textbf{real}        & \textbf{sketch}      & \textbf{Avg}         \\
\midrule
ERM                  & 58.1 $\pm$ 0.3       & 18.8 $\pm$ 0.3       & 46.7 $\pm$ 0.3       & 12.2 $\pm$ 0.4       & 59.6 $\pm$ 0.1       & 49.8 $\pm$ 0.4       & 40.9                 \\
IRM                  & 48.5 $\pm$ 2.8       & 15.0 $\pm$ 1.5       & 38.3 $\pm$ 4.3       & 10.9 $\pm$ 0.5       & 48.2 $\pm$ 5.2       & 42.3 $\pm$ 3.1       & 33.9                 \\
GroupDRO             & 47.2 $\pm$ 0.5       & 17.5 $\pm$ 0.4       & 33.8 $\pm$ 0.5       & 9.3 $\pm$ 0.3        & 51.6 $\pm$ 0.4       & 40.1 $\pm$ 0.6       & 33.3                 \\
Mixup                & 55.7 $\pm$ 0.3       & 18.5 $\pm$ 0.5       & 44.3 $\pm$ 0.5       & 12.5 $\pm$ 0.4       & 55.8 $\pm$ 0.3       & 48.2 $\pm$ 0.5       & 39.2                 \\
MLDG                 & 59.1 $\pm$ 0.2       & 19.1 $\pm$ 0.3       & 45.8 $\pm$ 0.7       & 13.4 $\pm$ 0.3       & 59.6 $\pm$ 0.2       & 50.2 $\pm$ 0.4       & 41.2                 \\
CORAL                & 59.2 $\pm$ 0.1       & 19.7 $\pm$ 0.2       & 46.6 $\pm$ 0.3       & 13.4 $\pm$ 0.4       & 59.8 $\pm$ 0.2       & 50.1 $\pm$ 0.6       & 41.5                 \\
MMD                  & 32.1 $\pm$ 13.3      & 11.0 $\pm$ 4.6       & 26.8 $\pm$ 11.3      & 8.7 $\pm$ 2.1        & 32.7 $\pm$ 13.8      & 28.9 $\pm$ 11.9      & 23.4                 \\
DANN                 & 53.1 $\pm$ 0.2       & 18.3 $\pm$ 0.1       & 44.2 $\pm$ 0.7       & 11.8 $\pm$ 0.1       & 55.5 $\pm$ 0.4       & 46.8 $\pm$ 0.6       & 38.3                 \\
CDANN                & 54.6 $\pm$ 0.4       & 17.3 $\pm$ 0.1       & 43.7 $\pm$ 0.9       & 12.1 $\pm$ 0.7       & 56.2 $\pm$ 0.4       & 45.9 $\pm$ 0.5       & 38.3                 \\
MTL                  & 57.9 $\pm$ 0.5       & 18.5 $\pm$ 0.4       & 46.0 $\pm$ 0.1       & 12.5 $\pm$ 0.1       & 59.5 $\pm$ 0.3       & 49.2 $\pm$ 0.1       & 40.6                 \\
SagNet               & 57.7 $\pm$ 0.3       & 19.0 $\pm$ 0.2       & 45.3 $\pm$ 0.3       & 12.7 $\pm$ 0.5       & 58.1 $\pm$ 0.5       & 48.8 $\pm$ 0.2       & 40.3                 \\
ARM                  & 49.7 $\pm$ 0.3       & 16.3 $\pm$ 0.5       & 40.9 $\pm$ 1.1       & 9.4 $\pm$ 0.1        & 53.4 $\pm$ 0.4       & 43.5 $\pm$ 0.4       & 35.5                 \\
VREx                 & 47.3 $\pm$ 3.5       & 16.0 $\pm$ 1.5       & 35.8 $\pm$ 4.6       & 10.9 $\pm$ 0.3       & 49.6 $\pm$ 4.9       & 42.0 $\pm$ 3.0       & 33.6                 \\
RSC                  & 55.0 $\pm$ 1.2       & 18.3 $\pm$ 0.5       & 44.4 $\pm$ 0.6       & 12.2 $\pm$ 0.2       & 55.7 $\pm$ 0.7       & 47.8 $\pm$ 0.9       & 38.9                 \\
MetaReg              & 59.8                 & \textbf{25.6}                 & 50.2                 & 11.5                 & 64.6                 & 50.1                 & 43.6                 \\
DMG                  & 65.2                 & 22.2                 & 50.0                 & 15.7                 & 59.6                 & 49.0                 & 43.6                 \\
SWAD                 & 66.0 $\pm$ 0.1       & 22.4 $\pm$ 0.3       & 53.5 $\pm$ 0.1       & \textbf{16.1} $\pm$ 0.2       & 65.8 $\pm$ 0.4       & 55.5 $\pm$ 0.3       & 46.5                 \\
\midrule
DART \small w/o SWAD & 65.9                 & 21.9                 & 52.6                 & 15.1                 & 64.9                 & 54.3                 & 45.8                 \\
DART \small w/ SWAD  & \textbf{66.5}                 & 22.8                 & \textbf{54.2}                 & \textbf{16.1}                 & \textbf{67.3}                 & \textbf{56.3}                 & \textbf{47.2}                 \\
\bottomrule
\label{tabular:domain}
\end{tabular}}
\end{center}